\documentclass[sigconf,noacm, 10pt]{acmart}

\AtBeginDocument{%
  \providecommand\BibTeX{{%
    \normalfont B\kern-0.5em{\scshape i\kern-0.25em b}\kern-0.8em\TeX}}}

\acmJournal{TOG}
\acmVolume{37}
\acmNumber{4}
\acmArticle{111}
\acmMonth{8}

\usepackage{algorithmic}
\usepackage{graphicx}
\usepackage{textcomp}
\usepackage{xcolor}

\usepackage[ruled, vlined, noend]{algorithm2e}
\usepackage{multicol}

\usepackage{amsmath}
\usepackage{float}
\usepackage{hyperref}
\usepackage{graphicx}
\usepackage{caption}
\usepackage{comment}
\usepackage[nameinlink]{cleveref}
\usepackage{subfig}
\usepackage{soul}
\usepackage{caption}

\renewcommand\footnotetextcopyrightpermission[1]{}

\makeatletter
\def\@copyrightspace{\relax}
\makeatother
\setcopyright{none} 

\begin{document}

\title{Rosko: Row Skipping Outer Products for Sparse Matrix Multiplication Kernels
}


\author{Vikas Natesh}
\affiliation{%
  \institution{Harvard University}
  \city{Cambridge, MA}
  \country{USA}}
\email{vnatesh@g.harvard.edu}

\author{Andrew Sabot}
\affiliation{%
  \institution{Harvard University}
  \city{Cambridge, MA}
  \country{USA}}
\email{asabot@g.harvard.edu}

\author{H. T. Kung}
\affiliation{%
  \institution{Harvard University}
  \city{Cambridge, MA}
  \country{USA}}
\email{kung@harvard.edu}

\author{Mark Ting}
\affiliation{%
  \institution{Harvard University}
  \city{Cambridge, MA}
  \country{USA}}
\email{weiteting@g.harvard.edu}

\renewcommand{\shortauthors}{Trovato and Tobin, et al.}

\begin{abstract}
We propose Rosko---\underline{ro}w \underline{sk}ipping \underline{o}uter products---for deriving sparse matrix multiplication (SpMM) kernels in reducing computation and memory access requirements of deep neural networks (DNNs).
Rosko allows skipping of entire row computations during program execution with low sparsity-management overheads. 
We analytically derive sparse CPU kernels that adapt to given hardware characteristics to effectively utilize processor cores and minimize data movement without the need for auto-tuning or search space exploration. 
Rosko can be integrated with other outer product scheduling methods, allowing them to leverage row skipping by using Rosko's packing format to skip unnecessary computation.

Rosko kernels outperform existing auto-tuning and search-based solutions as well as state-of-the-art vendor-optimized libraries on real hardware across a variety of neural network workloads. 
For matrices with sparsities ranging from 65\% to 99.8\% typically found in machine learning, Rosko kernels achieve up to a 6.5x runtime reduction on Intel and ARM CPUs.
\end{abstract}

\maketitle
\pagestyle{plain} 

\section{Introduction}

Matrix multiplication (MM) is a fundamental computation in deep learning \cite{transformer_contractions}.
For example, each convolution layer within a convolutional neural network (CNN) can be represented as an MM between the inputs and the weights of a layer (see e.g., \cite{warden2015why, mcdanel2019full}).
These multiplications often involve large weight matrices with many elements.
One common approach to accelerating CNNs is leveraging sparsity to avoid unnecessary computations, or \textit{zero skipping}, and reduce the total amount of computations with zero operands performed.

We can typically increase the sparsity, or percentage of zeros, by pruning model weights of small magnitudes using iterative pruning.
Pruning can be applied at the level of individual weight elements (unstructured) or to groups of elements (structured) at varying granularities (see, e.g., \cite{anwar2017structured}).
These groups may be filters, channels, or entire layers of a CNN.
Structured pruning approaches have used LASSO or $\ell_1$ channel selection (see, e.g., \cite{he2015deep, liu2017learning}) to determine pruning targets (entire input channels, filters, layers, etc).
Since structured sparsity yields predictable nonzero distributions in weight matrices, it is amenable to software and/or hardware acceleration \cite{zhou2021learning} for skipping unnecessary computation.   
\begin{figure}
\centering
    \subfloat{\label{fig:rosko_intel}{\includegraphics[width=0.5\linewidth]{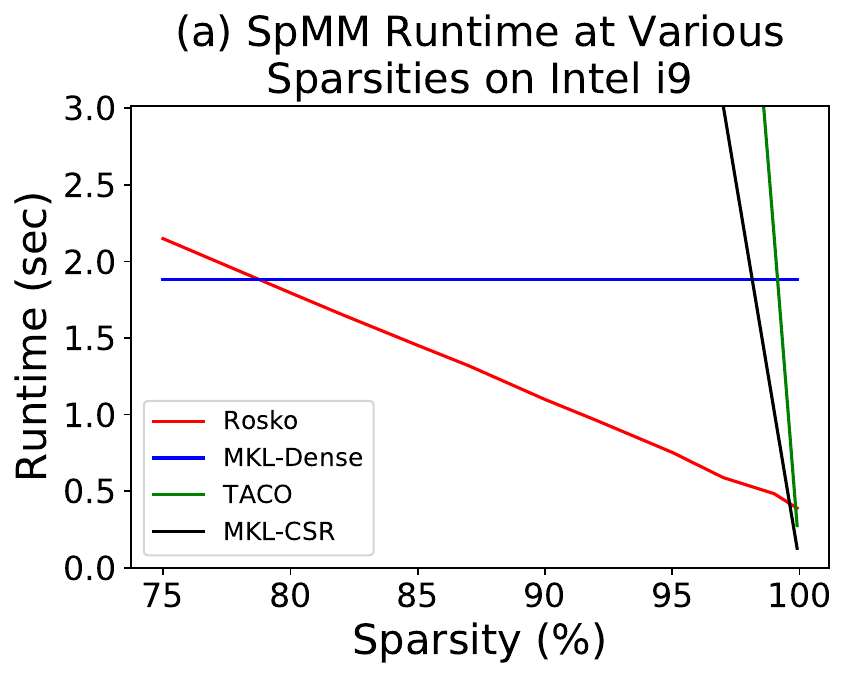}}}
    \subfloat{\label{fig:rosko_arm}{\includegraphics[width=0.5\linewidth]{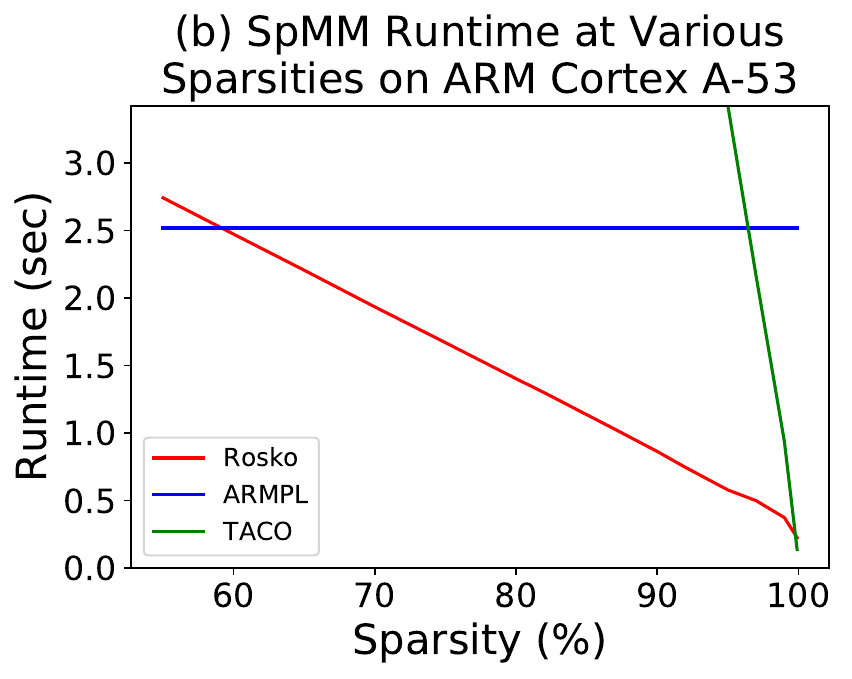}}}
\caption{
    Wall-clock SpMM runtime for Rosko kernels compared to sparse and dense libraries.
    Rosko outperforms dense MM and SpMM kernels from (a) 75\% to 99.5\% sparsity on Intel and (b) 58\% to 99.5\% sparsity on ARM CPUs.
    This figure is a preview of key results from \Cref{fig:intelperf,fig:armperf}. 
    An explanation for the larger sparsity range for ARM is given in \Cref{sec:validate}.
    }
\label{fig:intro}
\end{figure}

On the other hand, unstructured pruning is able to achieve a relatively high accuracy (see, e.g., \cite{han2015compression}) compared to structured pruning.
However, leveraging sparsity resulting from unstructured pruning is challenging: non-zero values may be arbitrarily distributed and must be addressed individually \cite{termquant, wang2021sparsednn}.
Vendor libraries that process sparse data formats, such as compressed sparse row (CSR) or compressed sparse column (CSC) formats \cite{dongarracrs,dongarraccs, intelmkl} may incur significant overheads due to indexing and irregular memory accesses \cite{narang2017blocksparse}.
Auto-tuning software, tensor compilers, and search-space exploration methods including TVM \cite{tvm}, TACO \cite{taco}, and FeatGraph \cite{featgraph} attempt to find high-performance schedules and tiling parameters for sparse computations. 
However, such methods do not necessarily utilize on-chip resources (caches, cores) to minimize data movement and often exhibit poor performance on multi-core systems.

We propose Rosko, a library that uses the outer product formulation in leveraging unstructured sparsity during sparse-matrix-times-dense-matrix multiplication (SpMM) with low sparsity management overhead.
Specifically, Rosko skips entire rows of computation corresponding to zero entries in the outer product column input, which result from zero-valued individual weights.

Rosko kernels perform sparse outer product computations with dense packed inputs.
Consequently, Rosko can work with existing scheduling approaches such as CAKE \cite{cake} or Goto's algorithm \cite{gotomain}, which reduce memory accesses for dense MM via increasing data reuse.
We introduce a novel sparsity-aware tiling algorithm that controls sparse tile size and shape to maximize arithmetic intensity given available on-chip memory, DRAM bandwidth, number of cores, and matrix sparsity (\%).
Our method is analytical and can efficiently tile SpMM problems with unstructured sparsity without auto-tuning or searching for tile sizes.
Although we demonstrate Rosko through SpMM kernels, Rosko's methodology is generally applicable. 
For example, Rosko can be applied to outer product-based methods for sparse direct convolution or tensor contractions. 
\Cref{sec:validate} empirically shows performance improvements from row skipping and better data reuse through tiling.

In evaluating Rosko's performance, we compare to vendor-optimized SpMM and dense MM CPU kernels.
We also compare to state-of-the-art compiler and auto-tuning approaches for sparse kernels in machine learning.
Existing SpMM kernels (e.g., MKL-CSR) underperform at lower sparsities due to sparsity management overhead (e.g., from indexing nonzeros) but excel at very high sparsities, as seen in \Cref{fig:intro}.
Meanwhile, dense MM kernels, such as \cite{intelmkl, xianyi2012openblas}, demonstrate consistent performance thanks to data streaming and reuse from algorithms such as Goto and CAKE, but do not benefit from increased sparsity.
Rosko aims to address the performance gap between sparse and dense libraries by extending outer product methods to skip zero computations and leveraging data streaming from libraries such as CAKE.
\Cref{fig:intro} is a visualization of Rosko's runtime improvements over wide ranges of sparsities.
{
The following summarizes this paper's contributions:
\begin{itemize}
    \item Rosko algorithm for efficient SpMM computation using row skipping outer products with low overhead in sparsity management from Rosko packing and sparse tiling (\Cref{sec:rosko_scheme}).
    Rosko is designed for machine learning, but other sparse applications such as those in scientific computing can also benefit from row skipping.    
    \item Efficient library implementations of Rosko kernels, improving performance over existing sparse and dense MM libraries on real CPU hardware for wide ranges of sparsities (\Cref{sec:perf_eval}).
    \item Extensive experiments demonstrating Rosko outperforms state-of-the-art vendor-optimized, compiler, and auto-tuning solutions for SpMM, in terms of computation throughput and memory bandwidth, across different neural network workloads and CPU platforms (\Cref{sec:perf_eval}).
\end{itemize}
}

\begin{figure}
    \centering
    \includegraphics[width=\linewidth]{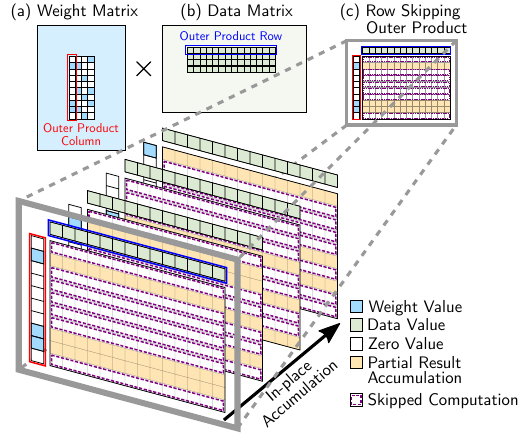}
    \caption{
    Overview of Rosko and sparse outer product MM.
    (a) Sparse weight matrix columns are multiplied with (b) dense data matrix rows using (c) row skipping outer products.
    Multiple outer products accumulate in-place during an MM.
    }
    \label{fig:approach_overview}
\end{figure}

{
The rest of this paper is organized as follows. We provide background and motivations for Rosko by examining related SpMM methods in \Cref{sec:background}.
\Cref{sec:novelty} overviews outer product row skipping and elaborates its motivations. 
\Cref{sec:rosko_scheme} details Rosko's design implementation with sparse block shaping.
We evaluate Rosko in \Cref{sec:perf_eval} and discuss the implications and further applications of Rosko in \Cref{sec:discussion}.
}

\begin{center}
\begin{table}
 \caption{Terminology Used Throughout the Paper}
  \label{tab:terms}
 \centering
 \scalebox{0.8}{
\begin{tabular}{ r l }
\hline\hline 
Term & Description \\
\hline 
{MAC} & {multiply-accumulate operation} \\
MM & matrix multiplication\\ 
nnz & number of nonzeros\\
Rosko & Row skipping outer product \\
Row skipping & skipping a row of computation within an outer product\\
SIMD & single instruction, multiple data\\
SpMM & sparse-matrix-times-dense-matrix multiplication \\
 Tile & small submatrix processed by a core or SIMD registers\\
\hline 
\end{tabular}
}
\end{table}
\end{center}

\section{Background and Related Work}
\label{sec:background}
In this section, we introduce some background and related works to motivate Rosko.
Rosko kernels fit into a sparsity regime not targeted by existing dense and sparse MM libraries.
They leverage outer products, block scheduling, and sparsity-aware tiling to perform efficient SpMM computations with minimal memory accesses.

\subsection{Outer Products for Dense MM}
\label{sec:outer_product_mm}
Outer product-based MM can attain a higher arithmetic intensity, or ratio of computation throughput to off-chip memory bandwidth, than inner product-based MM via efficient data streaming.
A higher {arithmetic intensity} enables more effective use of limited memory bandwidth.

\begin{figure}[h]
    \centering
    \includegraphics[width=\linewidth]{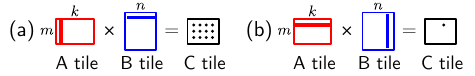}
    \caption{
    (a) Outer product and (b) inner product tile MM.
    The dots in $C$ represent elements computable with the highlighted columns and rows of the red $A$ and blue $B$ tiles.
    Outer products compute partial results for a $C$ tile using a column vector of the $A$ tile and the corresponding row vector of the $B$ tile.
    Inner products fully accumulate a single $C$ element using a row vector of the $A$ tile and column vector of the $B$ tile.
    }
    \label{fig:in_vs_out}
\end{figure}

We can divide a matrix multiplication, $C = A \times B$, of two matrices, $A$ and $B$, into smaller sub-problems.
Each sub-problem contributes to the computation of an $m \times n$ tile of $C$.
Given a value $k$, we can tile $A$ and $B$ into $m \times k$ and $k \times n$ submatrices, respectively, that compute partial $C$ tiles. 
Computation of a single $C$ tile can be viewed as a smaller, ``complete'' MM, and can be performed via inner or outer products, as seen in \Cref{fig:in_vs_out}.

Maximizing arithmetic intensity for a full MM requires maximizing arithmetic intensity of each tile.
Using inner products, streaming an additional $k$ values (e.g., another full row of the $A$ tile) enables only another $k$ MAC operations (in computing a single element of $C$).
This limits tile {arithmetic  intensity} to $\frac{k}{k} = 1$.
Using an outer product approach, we can increase arithmetic intensity by prioritizing keeping more values of $C$ in memory.
Streaming an additional $m + n$ values ($m$ per column of the $A$ tile and $n$ per row of the $B$ tile) then enables an additional $mn$ MAC operations (in computing partial values to accumulate for $mn$ elements).
Consequently, tile {arithmetic  intensity} is increased to $\frac{mn}{m + n}$.
Rosko uses outer products to increase the arithmetic intensity of tiled-SpMM.

Outer products are a widespread technique in many vendor libraries for dense MM \cite{intelmkl, armpl, armcl, BLIS, gotomain}.
Prior works such as \cite{pal2018outerspace, zhang2020sparch, wang2021dualside} also use outer products to accelerate SpMM directly in hardware.
However, these works only report simulated results or require hardware modification.
In contrast, Rosko utilizes outer products to yield a high-performance SpMM library directly on existing real CPUs (Intel, ARM, AMD, etc). 
We elaborate on why outer products are advantageous for handling sparsity in the \Cref{sec:novelty}.

\subsection{Block Scheduling For Better Data Reuse}
\label{sec:cake_sched}

Existing outer product-based scheduling algorithms for dense MM such as CAKE \cite{cake} and Goto's algorithm \cite{gotomain, intelmkl, BLIS} partition the MM and schedule each tile according to different objectives. 
For example, Intel MKL \cite{intelmkl} attempts to maximize throughput by tiling the MM to utilize all available DRAM bandwidth.
On the other hand, CAKE \cite{cake} tiles the MM based on available on-chip memory to minimize DRAM bandwidth usage and attain high throughput.
Like CAKE, Rosko sizes each tile to available on-chip memory and schedules tiles such that they are computed in the $K$-dimension first, followed by $M$ and $N$ dimensions.
This allows us to compute a tile of $C$ entirely in on-chip memory while accumulating partial result tiles in the $K$-dimension before writing the final result to off-chip memory. 
By packing sparse matrix tiles into densely packed computations, Rosko can leverage outer product scheduling algorithms such as CAKE and Goto.

While CAKE's tile shaping is suitable for low-bandwidth, dense MM on multiple cores, it fails to properly utilize system resources for SpMM operations.
SpMM is inherently bandwidth-bound and as matrix sparsity increases, DRAM bandwidth usage must increase to keep the cores busy processing nonzero computations \cite{bwbound}.
In \Cref{sec:sparsity-Aware_block_shaping}, we introduce a sparsity-aware tiling scheme that uses minimal DRAM bandwidth when sparsity is modest (e.g., 65\%-90\%) and increases DRAM bandwidth usage at higher sparsities ($> 90\%$). 

\subsection{Related SpMM Methods on CPUs}

\textbf{Vendor-optimized SpMM}: Current libraries only improve performance over dense MM for extremely high sparsities, as shown in \Cref{fig:intro}.
For example, Intel oneMKL's CSR-based SpMM routine performs poorly compared to dense MM libraries for sparsities below 99\%.
\cite{gale2020sparse} identified a similar sparsity regime (between those targeted by dense and sparse libraries) where performance can be improved on GPUs.\\
\textbf{Auto-tuning and Compiler Approaches}: Sparse tensor compilers such as TACO \cite{taco}, TVM \cite{tvm}, and FeatGraph \cite{featgraph} attempt to automatically generate sparse codes given a description of the program in a tensor intermediate representation (IR).
Tensor compilers often perform a heuristic search of possible schedules and/or auto-tuning to identify optimal tile sizes after loop splitting transformations.
However, the generated schedules do not directly account for available on-chip (caches, cores) and off-chip (DRAM bandwidth) resources \cite{tacosched}.
Searching may also introduce overhead in situations when sparsity changes dynamically (e.g., pruning weight matrices during training), due to repeated searches.
As a result, tensor compiler-generated code may not outperform vendor libraries in practice. 
Rosko may improve the runtime of tensor compiler code because Rosko's tiling parameters are derived analytically, without searching through possible schedules or tile sizes.

\section{Overview of Row Skipping Outer Products}
\label{sec:novelty}
In this section, we review the motivations for Rosko's proposed outer product row skipping approach, summarize its novelty, and provide an overview.

\begin{figure}
    \centering
    \includegraphics[width=\linewidth]{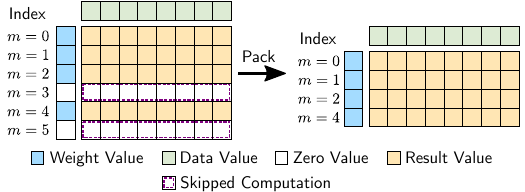}
    \caption{
    Example of a row skipping outer product between a sparse column (blue) and dense row (green) to obtain a result matrix (yellow).
    {White indicates zero values}, and purple outlines indicate skipped rows of computation.
    Here, row computations for indices $m=3$ and $m=5$ are skipped and omitted from the resulting packed computation.
    }
    \label{fig:rosko_scheme}
\end{figure}

\Cref{fig:approach_overview} illustrates the outer product SpMM, where $A$ is a sparse weight matrix, $B$ is a dense data matrix, and $C$ is the in-place accumulation results.
Each outer product is between a column vector $a$ of $A$ and the corresponding row vector $b$ of $B$.
Each result row of an outer product depends on only one element from $a$.
Thus, we can determine whether to skip a whole row of computation simply by checking if an element is zero, as shown in \Cref{fig:rosko_scheme}.
Intuitively, this yields a simple streaming scheme where only nonzero elements of $a$ need to be streamed in to be used with stationary elements of $b$.

We take advantage of this streaming scheme in outer product column sparsity: by removing row computations associated with a zero element of a column vector of $A$, that element will never need to be loaded.
Removing computation rows from an outer product also allows us to pack the sparse column inputs, as shown in \Cref{fig:rosko_scheme}.
This results in a simple packing scheme with low overhead during computation.


For any general-purpose SIMD architecture, row skipping outer products are simple to implement and do not require hardware modification. 
Unlike prior work \cite{yu2017scalpel}, Rosko kernels can leverage SIMD units even for non-contiguous nonzero weights.
Inputs to row skipping outer products need only a simple indexing scheme (described in \Cref{sec:packing}), so there is minimal unpacking and indexing overhead.
In summary, our novelty lies in realizing that:
\begin{itemize}
    \item Outer product column sparsity allows for skipping row  computations based on whether individual elements of a column input vector are nonzero.
    \item We can efficiently pack outer products with sparse column input vectors into dense matrices and perform tiling to increase data reuse, leverage the CPU's SIMD hardware, and utilize multiple cores. (\Cref{sec:packing}).
\end{itemize}

\section{Rosko System Design and Optimization}
\label{sec:rosko_scheme}

Rosko's implementation leverages outer product-based SpMM scheduling by packing sparse outer product columns into dense columns and maximizing data reuse via sparsity-aware tiling.
We elaborate on the design and benefits of each technique in the following subsections.

\begin{figure*}[h]
    \subfloat{\label{fig:reorder_a}{\includegraphics[width=0.33\textwidth]{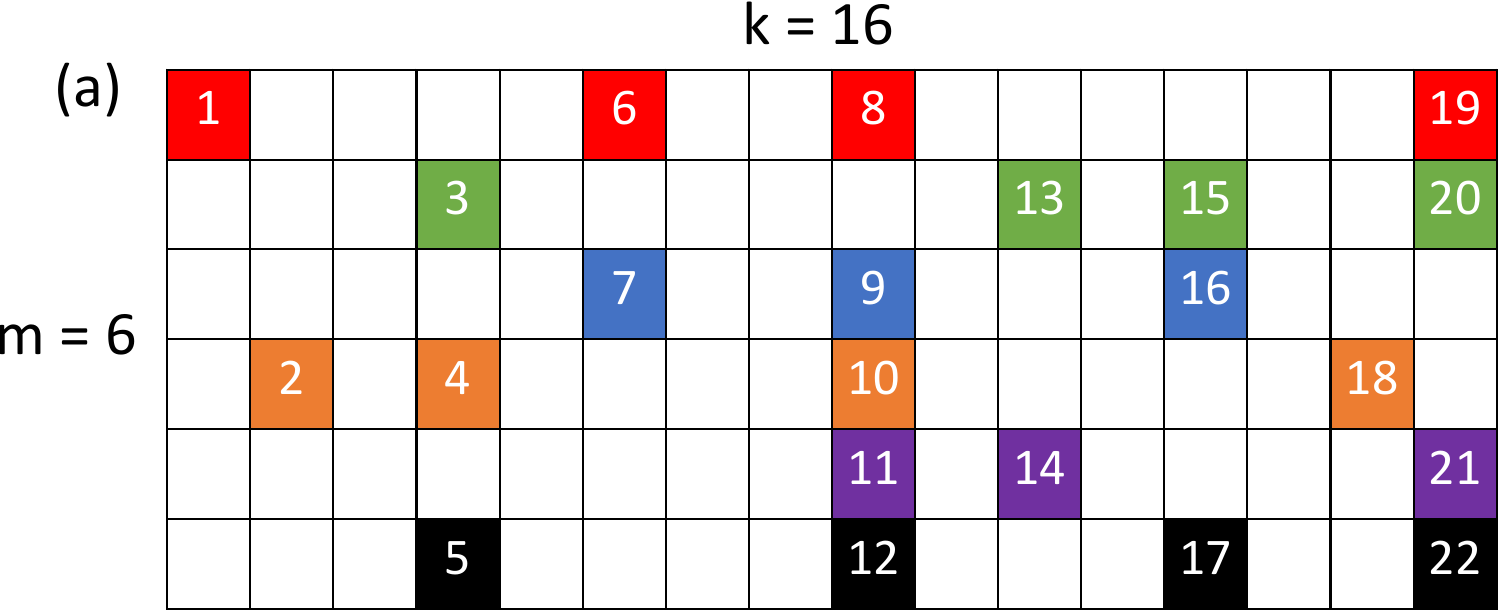}}}\hfill
    \subfloat{\label{fig:reorder_b}{\includegraphics[width=0.32\textwidth]{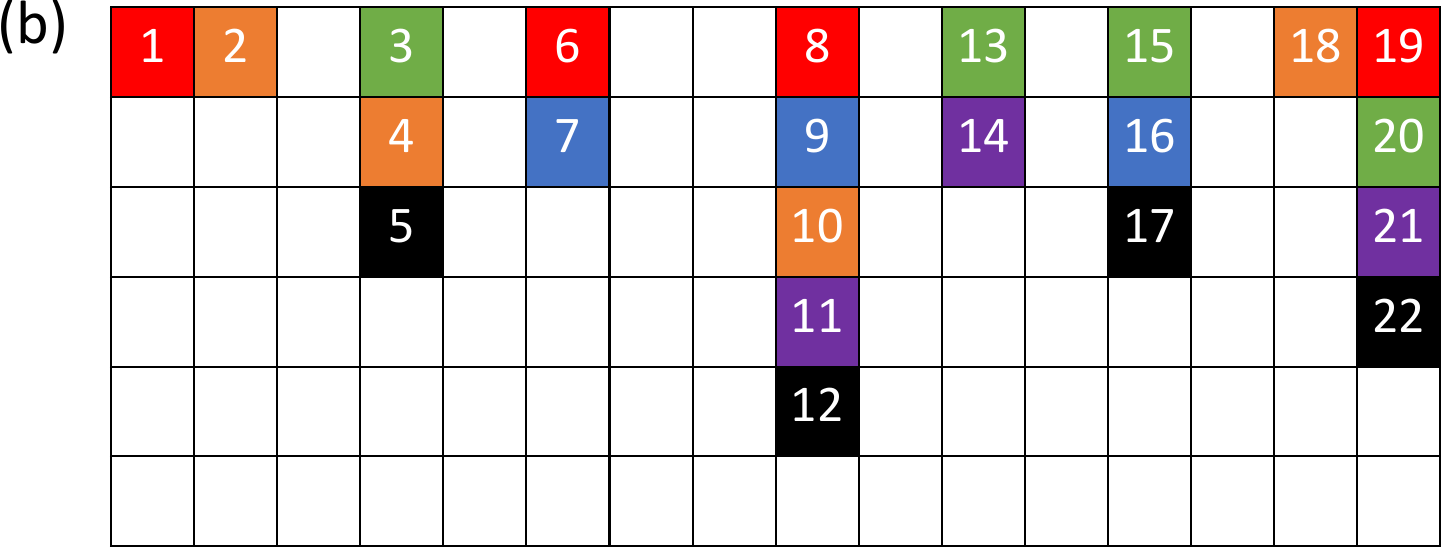}}}\hfill
    \subfloat{\label{fig:reorder_c}{\includegraphics[width=0.32\textwidth]{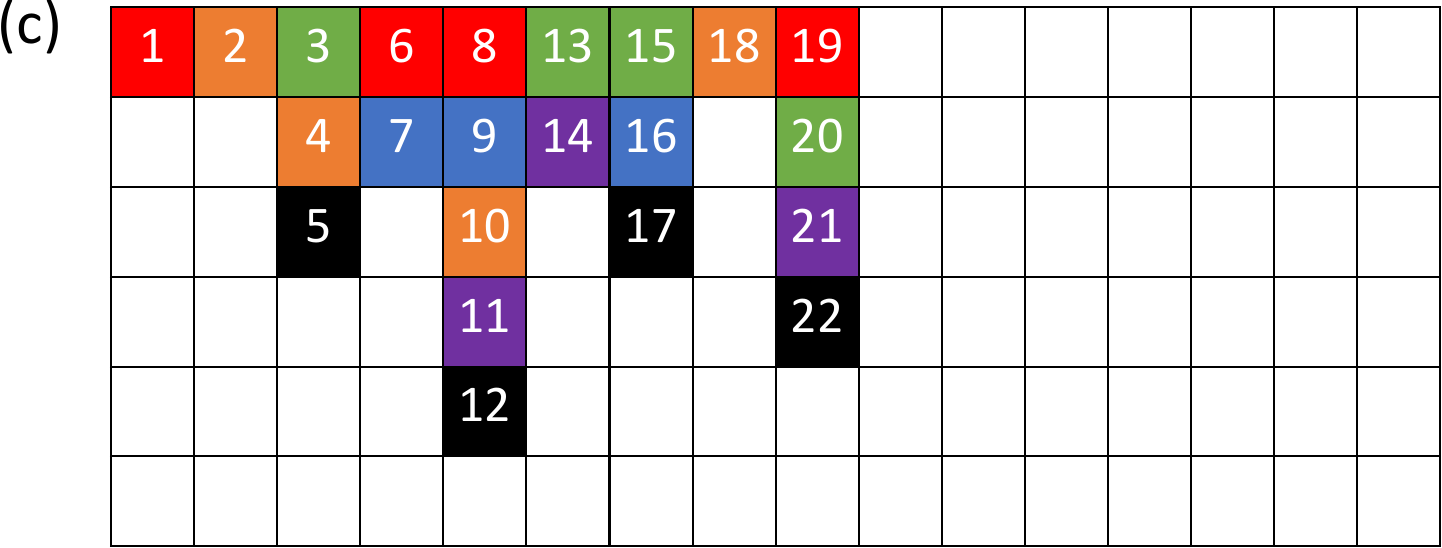}}}\hfill
\caption{
    {
	Visualizing row-skipped outer product packing.
	(a) $6\times16$ outer product tile with random sparsity.
	Colored squares are nonzero values; all nonzeros in the same row have the same color.
	(b) We pack nonzeros into contiguous columns.
	(c) Pack columns containing nonzero values into a contiguous buffer with values ordered 1,2,...,22.
    }
} \label{fig:reorder}
\end{figure*}

\subsection{Skipping Rows Within Outer Products via Dense Rosko Packing}
\label{sec:packing}
Rosko kernels perform outer products between sparse $A$ tiles and dense $B$ tiles (\Cref{fig:in_vs_out}), skipping unnecessary operations and memory accesses by taking advantage of column input sparsity, as depicted in \Cref{fig:rosko_scheme}.

Rosko packs nonzero values of $A$ columns into contiguous buffers and employs indexing arrays for element accesses during SpMM (demonstrated in \Cref{algo:pack} and \Cref{algo:rosko}) 
The original $m$-dimension index of each nonzero value is stored in an index array ($loc_m$).
Using this index array, the result row of an outer product is matched to the correct row in result tile $C$.
The counts of nonzero elements in each outer product column is stored in another array ($nnz$), which allows us to fix the loop bound of each outer product (i.e., number of nonzeros in the column) at runtime. 
Analogous to Rosko, conventional inner-product-based compressed sparse row (CSR) formats for SpMM track column indices for each nonzero value and the number of nonzeros in each row.

\begin{figure}[ht]
\centering
    \subfloat{\label{fig:pack_perf}{\includegraphics[width=0.49\linewidth]{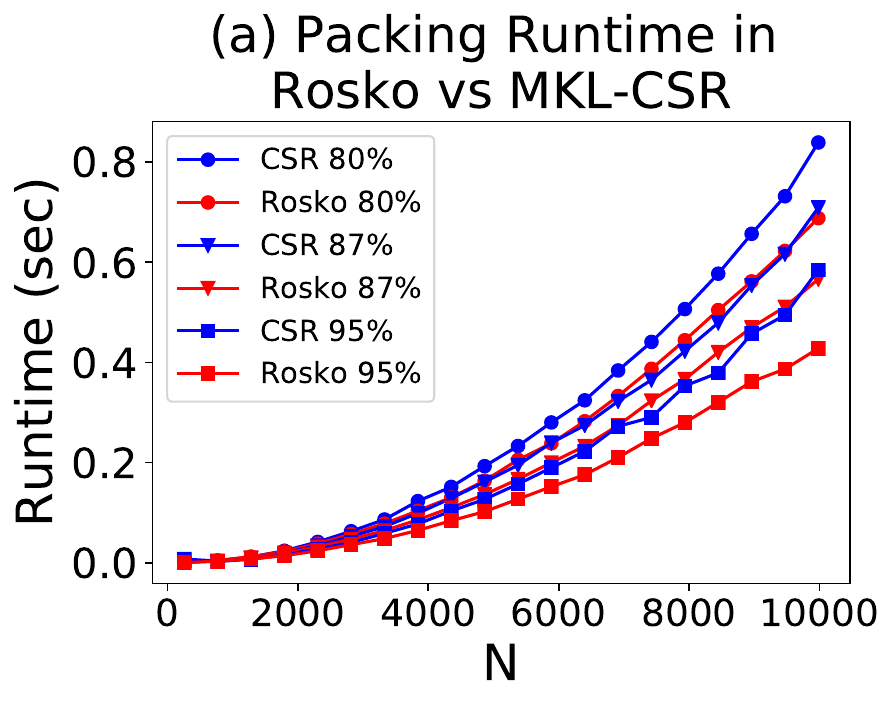}}}
    \subfloat{\label{fig:pack_dram}{\includegraphics[width=0.53\linewidth]{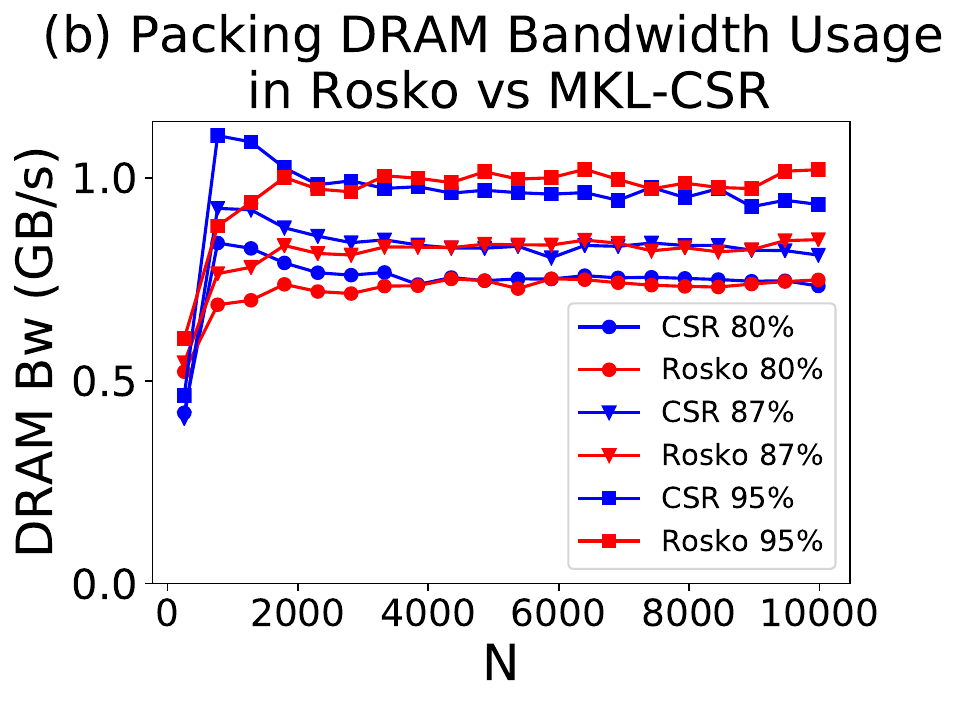}}}
\caption{
Single-core packing runtime and DRAM bandwidth when performing SpMM between two $N\times N$ matrices using MKL-CSR and Rosko. 
Rosko outperforms or matches the runtime of CSR with comparable DRAM bandwidth usage across several sparsity levels (denoted as \%) and matrix sizes, making it suitable for dynamic SpMM execution with on-demand packing.
} 
\label{fig:pack}
\end{figure}

To avoid packing overhead during DNN inference computations, sparse weight matrices for each layer can be pre-packed and stored.
However, applications that dynamically generate sparse matrices (e.g., pruning during DNN training and scientific simulations) require on-the-fly packing before sparse computations \cite{kolodziej2019suitesparse}.
A packing method that can fully utilize available DRAM bandwidth for the required memory copies is needed to reduce overhead in such settings.
We compare single-core performance of oneMKL's CSR packing format to Rosko's row skipping packing on an Intel i9 CPU (see \Cref{sec:perf_eval} for platform details).
The CSR format is widely supported by most vendor libraries and tensor compilers.
\Cref{fig:pack} shows Rosko's packing overhead is on-par with oneMKL's CSR format in terms of runtime while using a similar amount of DRAM bandwidth.
Hence, Rosko packing is a reasonable option for scenarios where sparse matrices are generated dynamically.

\small{
\begin{algorithm}
    \SetAlgoLined
    \SetInd{0.25em}{0.5em}
    \textbf{Input:} 
\begin{itemize}
    \item Sparse $A$ tile of size $M\times K$\\(Note: $A$ is accessed as a pointer in Algorithms 1 and 2.)
    
\end{itemize}

\textbf{Output:}
\begin{itemize}
    \item Packed sparse outer-product tile $A_p$
    \item $nnz$ array of number of nonzeros across $K$ cols of $A$
    \item $col_{ind}$ array of $A$ column indices
    \item $loc_m$ array of $m$-dim locations to write result rows of $C$
\end{itemize}
    
    $i_1 = 0$; $i_2 = 0$; \tcp{temp indices}
    \For{$k = 0$ \KwTo $K-1$}{
        $cols = 0$\;
        \For{$m = 0$ \KwTo $M-1$}{
            \uIf{$A[k + mK] \neq 0$}
                {$A_p[i_1] = A[k + mK]$\;
                $loc_{m}[i_1$++$] = m$\;
                $cols$++\;}
        }

        {
        \uIf{$cols \neq 0$}
            {$col_{ind}[i_2] = k$\;
            $nnz[i_2$++$] = cols$\;}
        }
    }

\caption{Rosko Packing and Indexing Arrays}
\label{algo:pack}
\end{algorithm}
}

\subsection{Sparsity-Aware Tiling with Rosko}
\label{sec:sparsity-Aware_block_shaping}

\begin{figure}
    \centering
    \includegraphics[width=\linewidth]{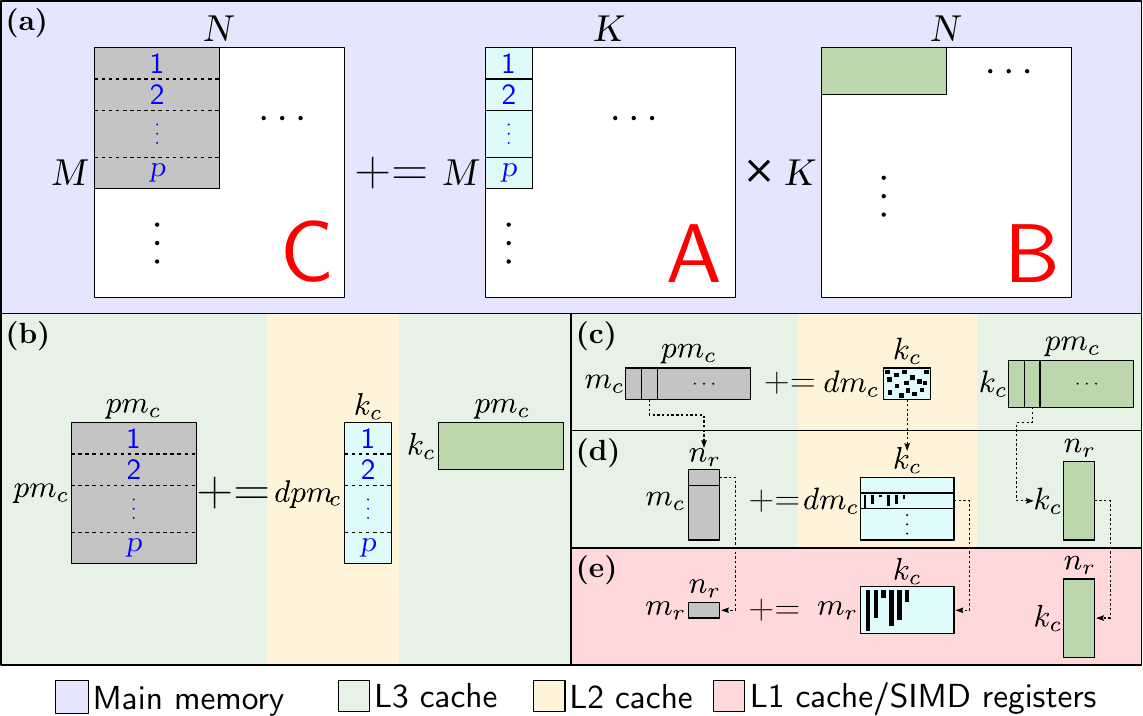}
    \caption{
    Rosko's tiling and scheduling in the CPU memory hierarchy. Each of the $p$ sparse sub-matrices from $A$ is reused in the L2 cache of a core while data from $B$ and partial results for $C$ are reused in the L3 cache. Boxes (c) and (d) show computation of a single $m_c \times pm_c$ sub-matrix of $C$ on a core with the unstructured sparsity of $A$ tile shown by the random squares in (c). Box (e) shows the row-skipping outer product with packed outer product columns of A (black lines). See \Cref{algo:rosko} for details.
    }
    \label{fig:rosko_tiling}
\end{figure}

Modern CPUs contain a multilevel memory hierarchy with local caches on each core, one or more shared caches between cores, and off-chip memory (e.g., DRAM).
Traditional multilevel tiling methods for dense MM are not optimal for SpMM: resource requirements (e.g., needed off-chip memory bandwidth) will vary based on input sparsity. 
Several existing SpMM methods rely on auto-tuning or searching for tile sizes to accomplish tiling and scheduling at multiple memory levels.
However, such methods do not directly manage system resources as part of their objective and may result in poor performance.
In this section, we show that Rosko yields an analytical tiling method that achieves high throughput for SpMM.
By analysis, we show that Rosko manages system resources such as DRAM bandwidth, on-chip memory, and CPU cores while taking into account input matrix sparsity to attain high throughput.

For clarity of analysis, we assume sparse matrices have uniform random sparsity, where each sparse tile of $A$ has roughly the same density $d$, the fraction of values that are nonzero.
\Cref{fig:rosko_tiling} shows Rosko's tiling on a CPU with a DRAM/L3/L2/L1 memory hierarchy and $p$ processor cores.
The $m_c$, $k_c$, $m_r$, and $n_r$ are cache-level tile dimension sizes for input matrices $A$, $B$, and $C$, which we will obtain by analysis.

In \Cref{fig:rosko_tiling}a, we read $A$ and $B$ sub-matrices from DRAM into the L3 cache, allowing us to compute a square $pm_c\times pm_c$ partial result tile of $C$.
Using a $K$-first block schedule \cite{cake}, we continue to load $A$ and $B$ sub-matrices in the $K$-dimension while accumulating the partial result tile in the L3 cache.
Partial results are returned to DRAM only after being reused for all $K$-dimension accumulations. 
Hence, the off-chip IO required to fully compute a $pm_c\times pm_c$ tile of $C$ is:
\[
IO = IO_{A}+IO_{B} = 3dpm_ck_c + pm_ck_c
\]
The factor of 3 accounts for additional IO due to Rosko indexing arrays which, in the worst case, each contain no more than $dm_ck_c$ nonzero values. 
To support this computation and buffer the partial result tile for $K$ accumulations, the required on-chip memory is:
\[
3dpm_ck_c + pm_ck_c + p^2m_c^2 \leq L3 \tag{1}
\] 
By setting $m_c=k_c$, we can solve for $m_c$ and $k_c$ such that the $A$, $B$, and $C$ tiles fit in the L3 cache.

In \Cref{fig:rosko_tiling}b, the $A$ sub-matrix is partitioned into $dpm_c\times kc$ tiles, and each tile is loaded into the L2 cache of a core.
The $k_c \times pm_c$ sub-matrix of $B$ is then broadcast from the L3 cache to each core, allowing the $p$ cores to compute separate $m_c\times pm_c$ partial result tiles of $C$ in parallel.
Each core performs $dm_ck_cpm_c$ MAC operations. Assuming a fixed single-core peak throughput $peak$, the time required to compute the $pm_c\times pm_c$ tile of $C$ in L3 is:
\[
T = \frac{dm_ck_cpm_c}{peak} 
\] 
Given the expected runtime and memory accesses required to compute a partial tile of $C$ in L3, the required off-chip memory bandwidth is:
\[
BW = \frac{IO}{T} = \frac{3dpm_ck_c + pm_ck_c}{dm_ck_cpm_c} \cdot peak 
= \frac{3d + 1}{dm_c} \cdot peak \tag{2}
\] 

SpMM computations are bandwidth-bound, so off-chip bandwidth requirements for Rosko-SpMM grows as density $d$ decreases.
This is expected because opportunities for data reuse decrease as the matrix becomes more sparse. 
The cores must be loaded with new data for computation more frequently, thereby increasing the required DRAM bandwidth.
With Rosko, we can attain high throughput by tiling the computation such that DRAM bandwidth usage increases as matrix density decreases, via equation (2).

At the level of the SIMD registers and L1 cache on a single core in \Cref{fig:rosko_tiling}e, $dm_r \times k_c$ sparse packed tiles of $A$ are multiplied with $k_c \times n_r$ dense tiles of $B$ while $m_r \times n_r$ partial result tiles of $C$ are held in the registers for subsequent accumulations ($dm_rn_rk_c$ MAC ops).
Rosko packing allows for a vectorized sparse outer product implementation.
Each packed nonzero value of $A$ is broadcast to a SIMD register and multiply-accumulates with a row of $B$ to produce a partial result row of $C$, as shown in \Cref{algo:rosko}.
We then choose $m_r$ and $n_r$ to maximize outer product arithmetic intensity.
From \Cref{sec:outer_product_mm}, we know dense outer product arithmetic intensity $\frac{mn}{m+n}$ is maximized when $m$ and $n$ are roughly equal and as large as possible.
Similarly, for Rosko outer products, the arithmetic intensity $\frac{dmnk_c}{dmk_c+nk_c}=\frac{dmn}{dm+n}$ is also maximized when $dm$ and $n$ are roughly equal.
In practice, outer product tile size is constrained by L1 cache capacity.
Thus, we choose $m_r$ and $n_r$ such that $A$, $B$, and $C$ tiles fit in L1:
\[
3dm_rk_c + k_cn_r + m_rn_r \leq L1 \tag{3}
\] 
The values of $m_r$ and $n_r$ may also be constrained by available SIMD register lengths.
For example, to use 256-bit Intel AVX2 registers with 32-bit floating point values, $n_r$ must be a multiple of 8.

\small{
\begin{algorithm}
    \SetAlgoLined
    \SetInd{0.25em}{0.5em}
    \textbf{Input:} 
\begin{itemize}
    \item Packed sparse tile $A_p$ of size $M\times K$
    \item $K\times N$ tile of $B$ and $M\times N$ tile of $C$
    \item $nnz$ array of number of nonzeros across $K$ cols of $A_p$
    \item $col_{ind}$ array of $A_p$ column indices
    \item $loc_m$ array of indices for writing each result row of $C$
\end{itemize}
\textbf{Output:}
\begin{itemize}
    \item Outer product result tile $C$
\end{itemize}
    \tcp{Load $M$ rows of $C$ into SIMD regs}
    \For{$m = 0$ \KwTo $M-1$}{
        $c[m] = \text{load}(C[m])$\;
    }
    \tcp{Perform outer product}
    \For{$k = 0$ \KwTo $K-1$}{
        \tcp{Skip remaining cols w/o nonzeros}
        \uIf{$nnz[k] == 0$}{break\;}
        \tcp{Load correct $B$ row into SIMD regs using $col_{ind}$} 
        $b = \text{load}(B[col_{ind}[k]])$\;
        \For{$m = 0$ \KwTo $nnz[k]$}{
            \tcp{Broadcast nonzero $A_p$ value to SIMD regs and advance $A_p$ pointer}
            {
            $a = \text{broadcast}(*(A_p$++$))$\;}
            \tcp{SIMD MAC operation with result written to correct row index}
			{$c[*loc_m] = \text{vmac}(a, b, c[*loc_m])$\;}
            \tcp{Advance $loc_m$ array pointer for next row}
			{$*loc_m$++\;}
        }
    }
    \tcp{Store $M$ rows of $C$ back to memory}
    \For{$m = 0$ \KwTo $M-1$}{
        $C[m] =$ store$(c[m])$\;
    }

\caption{Packed Row Skipping Outer Product}
\label{algo:rosko}
\end{algorithm}
}

\section{Performance Evaluation of Rosko on CPUs}
\label{sec:perf_eval}

We compare Rosko to state-of-the-art dense MM and SpMM methods on two CPU architectures (see \Cref{tab:cpus} and \Cref{sec:cpu_setup}) for various SpMM problems.
Rosko is implemented in C\texttt{++}, using Intel AVX2 and ARM Neon SIMD intrinsics.
When evaluating the performance of Rosko and SpMM libraries, we use computation throughput (GFLOPs/sec) as our metric, where the number of FLOPs is computed as $d \cdot MKN = \#nnz \cdot N$, where $d$ is the fraction of values that are nonzero in the $M\times K$ sparse matrix. 
For memory usage, we report DRAM bandwidth (GB/sec) and total DRAM IO (GB) as our metrics. 

In the subsequent sections, we evaluate Rosko on SpMM computations that arise during DNN inference and training.
Sparse matrices in DNNs have various shapes, sparsity patterns, and sparsity ranges\cite{kolodziej2019suitesparse}.
In \Cref{sec:validate}, we validate throughput and predicted DRAM bandwidth usage of Rosko on both CPUs for synthetic sparse matrices. 
Then, in \Cref{sec:bench}, we compare Rosko to competing dense MM and SpMM methods using sparse transformer weight matrices from the Deep Learning Matrix Collection (DMLC) benchmark \cite{dlmc}.
These matrices have unstructured sparsity ranging from 70\% to 98\% resulting from various pruning methods such as magnitude pruning and variational dropout.
In \Cref{sec:sim}, we show Rosko can improve end-to-end performance of DNN training and inference on 3 CNN models.
Finally, in \Cref{sec:featgraph}, we compare throughput of Rosko and FeatGraph on the Intel CPU for SpMM computation in graph convolutional networks (GCN).

\subsection{CPU Evaluation Setup}
\label{sec:cpu_setup}
We evaluate Rosko on the Intel Core i9-10900K and ARM Cortex-A53 CPUs.
Each CPU has unique architecture and system characteristics, allowing us to profile Rosko under different constraints in off-chip memory bandwidth, on-chip memory size, and on-chip memory bandwidth (\Cref{tab:cpus}).
The Cortex-A53 is a low-power mobile CPU with a relatively low peak DRAM bandwidth (2 GB/sec), on-chip memory size, and on-chip memory bandwidth.
The i9-10900K is a 10-core desktop CPU with high peak DRAM bandwidth (40 GB/sec) and a large on-chip memory.

In all experiments, we average our performance measurements over 50 trials.
We flush the cache between runs to avoid cache reuse across successive runs and measure the execution time of each individual run.
We also disable simultaneous multithreading (SMT) and dynamic voltage and frequency scaling (DVFS) on each system to maintain a consistent clock frequency and reduce variability across runs.

\begin{center}
\begin{table}
\normalsize
 \centering
  \scalebox{0.65}{
\begin{tabular} {c c c c c c c c}
\hline\hline 
CPU & L1 & L2 & L3 & DRAM & Cores & LLC BW & DRAM BW\\
\hline 
 Intel i9-10900K & 32 KiB & 256 KiB & 20 MiB & 32 GB & 10 & 225 GB/s & 40 GB/s\\ 
 ARM Cortex-A53 & 16 KiB & 512 KiB & N/A & 1 GB & 4 & 16 GB/s & 2 GB/s\\ 
\hline 
\end{tabular}
}
 \caption{CPUs Used in Rosko Evaluation}
\label{tab:cpus}
\end{table}
\end{center}

\subsection{Validating Rosko's Tiling Model}
\label{sec:validate}

To validate Rosko's performance and tiling model, we compare Rosko to competing dense MM and SpMM libraries using synthetic sparse matrices with various levels of random uniform sparsity.
In \Cref{sec:rosko_scheme}, we modelled the DRAM bandwdith and on-chip memory requirements of Rosko, as a result of outer products, sparsity-aware tiling, and block scheduling. 
Here, we profile Rosko SpMM on Intel and ARM CPUs against metrics of DRAM bandwidth usage and runtime to confirm whether Rosko accurately manages system resources as predicted by the tiling model.
We measure DRAM Bandwidth usage on Intel using the VTune Profiler \cite{VTune}. 
On the ARM CPU, we record DRAM accesses using the Linux perf tool \cite{perf} by monitoring the ARM PMU event counter for L2 cache refills from DRAM.  

The oneMKL library \cite{intelmkl} contains both dense MM and SpMM routines optimized for Intel CPUs.
Meanwhile, ARM provides two libraries for dense MM on Cortex-A CPUs. ARM Compute Library (ARMCL) \cite{armcl} contains several ML-specific kernels, including dense MM and convolution, while ARM Performance Library (ARMPL) \cite{armpl} contains BLAS routines for scientific computing.
We also compare Rosko to TACO, a compiler that allows the user to generate optimized tensor processing programs from scheduling primitives written in the TACO API. 
For example, one may use the TACO API to generate an SpMM schedule with optimal loop order, tiling (loop split), SIMD vectorization, and multi-core parallelization strategy. 
We evaluate the optimized TACO SpMM CPU schedule provided in \cite{taco_opt}, that performs all of the above transformations.
In addition, we use sparse matrices stored in the CSR-format for both TACO and MKL and do not include packing overhead in our measurements.

\begin{figure*}[t]
    \centering
    \subfloat{\label{fig:intelperf}{\includegraphics[width=0.24\textwidth]{figures/rosko_vs_intel_valid.pdf}}}
    \subfloat{\label{fig:armperf}{\includegraphics[width=0.24\textwidth]{figures/rosko_vs_arm_valid.pdf}}}
    \subfloat{\label{fig:intel_dram}{\includegraphics[width=0.25\textwidth]{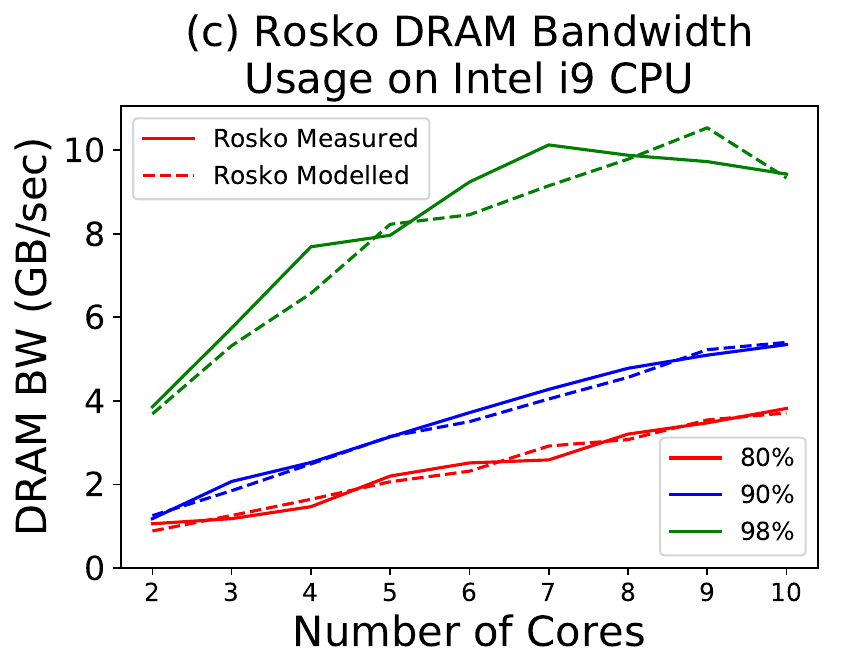}}}
    \subfloat{\label{fig:arm_dram}{\includegraphics[width=0.25\textwidth]{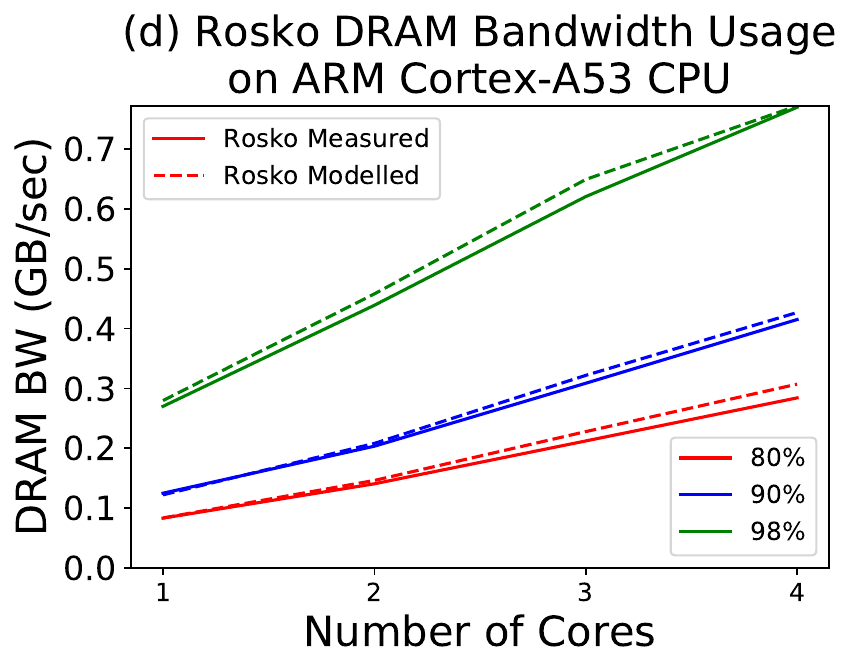}}}
\caption{
Validating Rosko's performance while varying sparsity. 
We run SpMM between $2000\times2000$ matrices on ARM and $10000\times10000$ matrices on Intel, measuring DRAM bandwidth usage and runtime.
Rosko uses sparsity-aware tiling to reduce DRAM bandwidth usage at lower sparsities and significantly improve throughput relative to ARMCL((a) and (b)) and MKL ((d) and (e)).
However, at high sparsities, Rosko's DRAM bandwidth usage is higher due to computation skipping, accounting for the bandwidth-bound nature of SpMM
} \label{fig:valid}
    \vspace{-4mm}
\end{figure*}

\Cref{fig:valid}(a) and (b) shows the runtime attained by different methods when performing SpMM between $10000\times10000$ matrices on Intel (a) and $2000\times2000$ matrices on ARM (b).
Rosko outperforms both dense MM and SpMM libraries in the sparsity range of 58\%-99.5\% on ARM and 75\%-99.5\% on Intel.
By directly minimizing DRAM bandwidth usage as a function of sparsity (equation 2), Rosko has greater performance gains on devices with constrained DRAM bandwidth.
Hence, Rosko improves runtime for a wider sparsity range on ARM (2 GB/sec) than Intel (40 GB/sec).

In \Cref{fig:valid}(c) and (d) we plot Rosko's modelled DRAM bandwidth requirements (equation 2) against measured bandwidth during SpMM. The predicted DRAM bandwidth usage matches the measured bandwidth to within 3\% error on both platforms.
As sparsity increases, Rosko achieves lower runtimes by utilizing more DRAM bandwidth.
Since entire rows of computations may be skipped, Rosko may complete the same size of a sparse MM in fewer operations compared to a dense MM kernel.

Results from the validation study show that Rosko's tiling model accurately predicts performance and bandwidth usage as a result of outer products, sparsity-aware tiling, and block scheduling.

\subsection{Performance on Transformer Benchmark For Sparsities from 70\%-98\%}
\label{sec:bench}

In this section, we profile Rosko on SpMM problems arising in transformer model training and inference.
Sparse matrices are selected from the DMLC benchmark \cite{dlmc}, which contains several pre-trained weight matrices from transformer model layers. 
Weight sparsity in the benchmark was generated using different pruning techniques during training e.g., random pruning, magnitude pruning, $\ell_0$ regularization, and variational dropout. 
For SpMM computation, we assume a batch size of 8 and sequence length of 256, thus fixing the $N$ dimension at $N=256*8=2048$ for $B$ and $C$ matrices.

In \Cref{fig:rosko_comparisons}, we plot throughput and speedup of Rosko compared to MKL and ARMPL dense MM, MKL-CSR SpMM, and TACO's CSR-based SpMM.
Rosko starts to outperform dense MM libraries when sparsity exceeds $\approx75\%$ on both CPUs.
Rosko also outperforms CSR-based SpMM methods in this sparsity range (70\% to 98\%), similar to \Cref{fig:valid}(a) and (b).

\Cref{fig:arm_google_bench} shows DRAM IO and bandwidth usage on the ARM Cortex-A53 as a function of the number of nonzeros in the SpMM.
Row skipping allows Rosko to require less total IO (\Cref{fig:arm_io}).
However, Rosko skips more computations relative to the reduction in memory accesses due to packing, i.e., runtime decreases more than total DRAM IO as sparsity increases.
Consequently, we observe Rosko uses more DRAM bandwidth for SpMM problems with fewer nonzeros, as shown in \Cref{fig:arm_bw}.
This is a favorable scenario in the sense that bandwidth usage increases only due to increased core utilization.
Thus, we can achieve lower runtimes when taking advantage of available DRAM bandwidth (\Cref{fig:arm_dram_tput}).


\subsection{Performance for End-to-End CNN Training}
\label{sec:sim}

We evaluate Rosko's performance versus ARMPL's dense MM on training three CNNs: VGG- 19 \cite{vggsimonyan2015deep}, MobileNetV2 \cite{howard2017mobilenets}, and ResNet-18 \cite{resnet} on CIFAR-10 \cite{Krizhevsky2009LearningML}.
CNN training can be decomposed as a series of matrix multiplications and data layout transformations.
One must compute output activations $O$ via $O=W\cdot A$ during the forward pass and activation gradients $\nabla A=\nabla O\cdot W^T$ and weight gradients $\nabla W=A^T \cdot \nabla O$ during the backward pass.
Since weight matrices become sparse with pruning while activations are treated as dense, the forward pass and activation gradient computation of the backward pass may be cast as SpMM operations for several training epochs.
By accelerating these two computations via SpMM, we can achieve a theoretical peak speedup of 3x over a dense MM library, assuming SpMM is applied during all epochs.
We implemented our own end-to-end training software for accurate comparison between dense MM and SpMM operations.

\begin{figure}[h]
    \centering
    \includegraphics[scale=0.3]{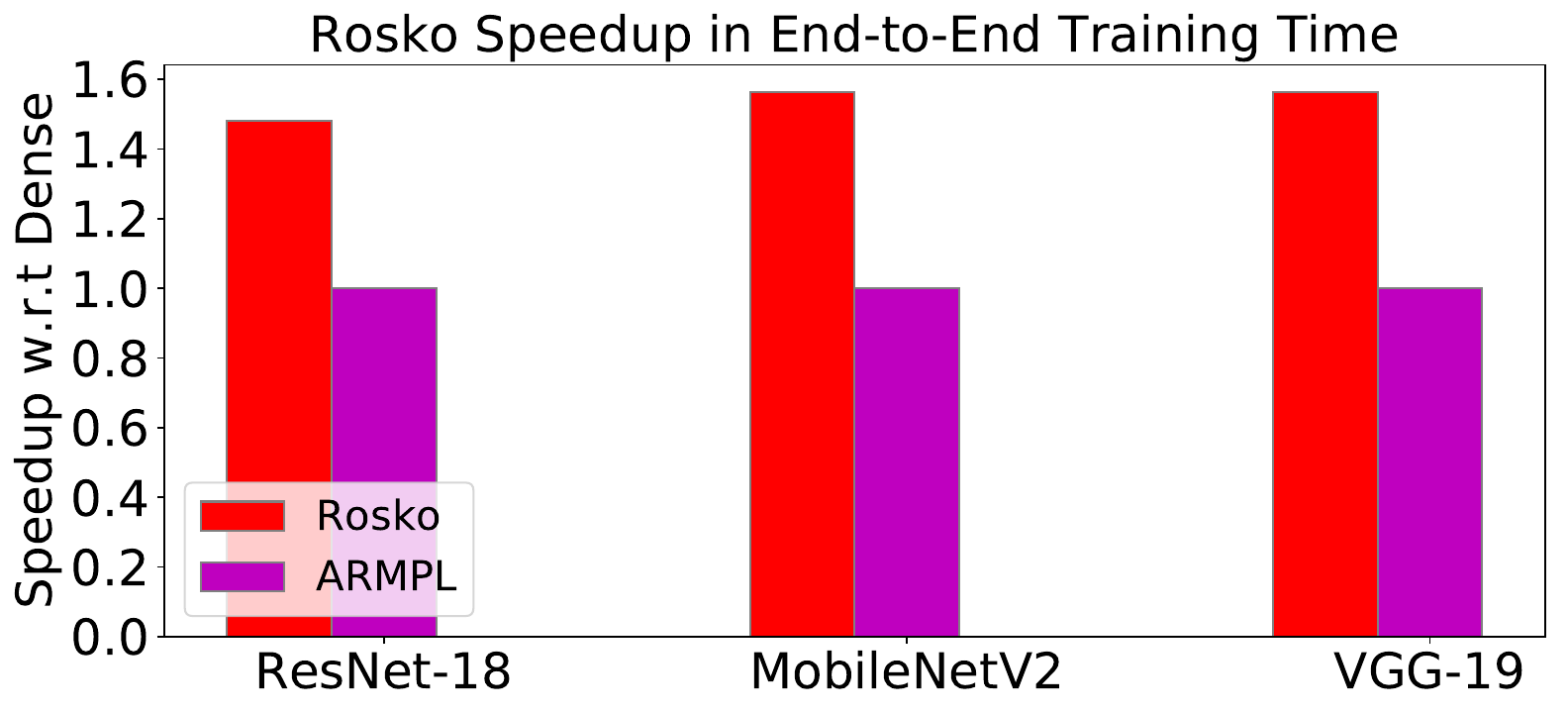}
    \caption{
    Rosko's speedup over ARMPL's dense MM during CNN training, using all 4 cores on ARM Cortex-A53. CNN models were trained on the CIFAR-10 dataset.
    Training with Rosko achieves a speedup of $\approx 1.5x$ accross networks.
    }
    \label{fig:arm_training}
\end{figure}
\begin{figure*}[ht]
    \subfloat{\label{fig:intel_tput_dlmc}{\includegraphics[width=0.25\textwidth]{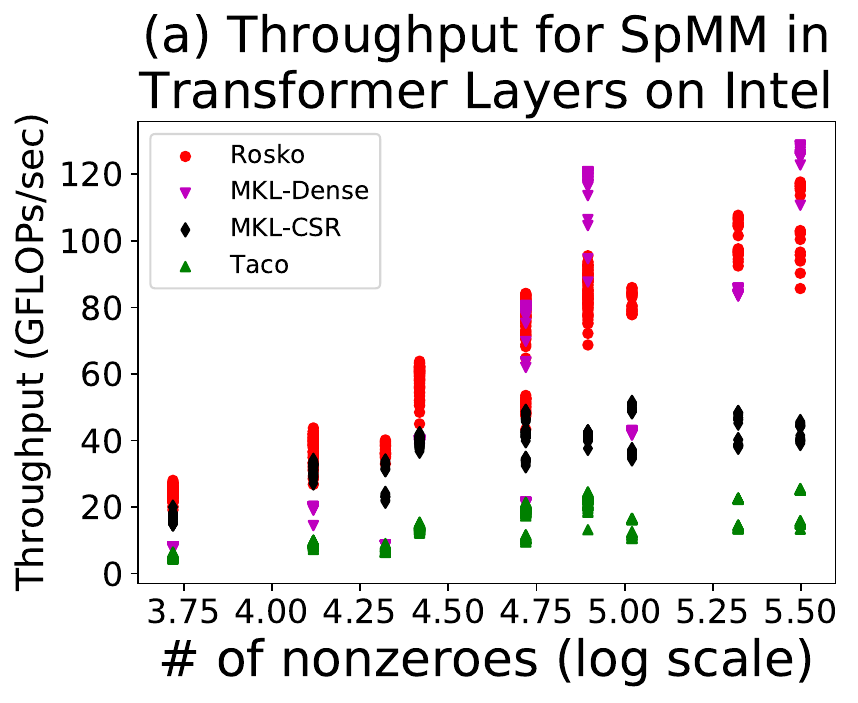}}}\hfill
    \subfloat{\label{fig:arm_tput_dlmc}{\includegraphics[width=0.25\textwidth]{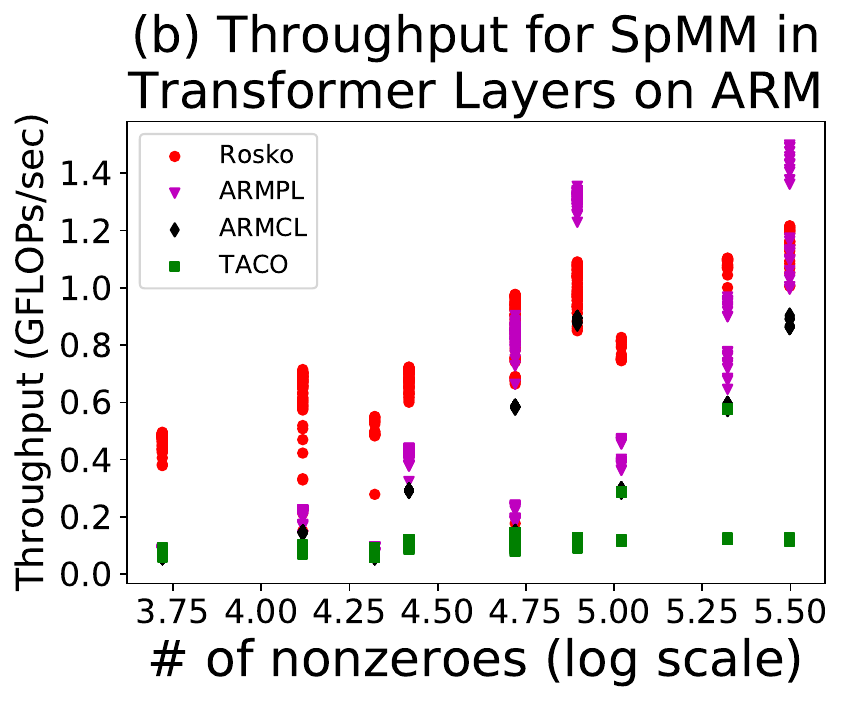}}}\hfill
    \subfloat{\label{fig:intel_tput}{\includegraphics[width=0.25\textwidth]{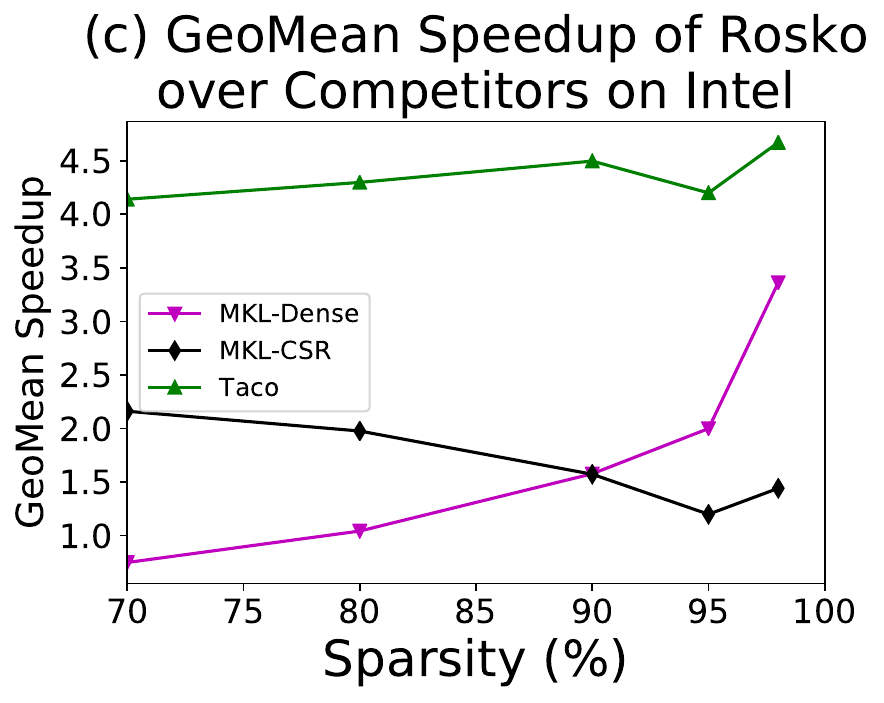}}}\hfill
    \subfloat{\label{fig:arm_tput}{\includegraphics[width=0.24\textwidth]{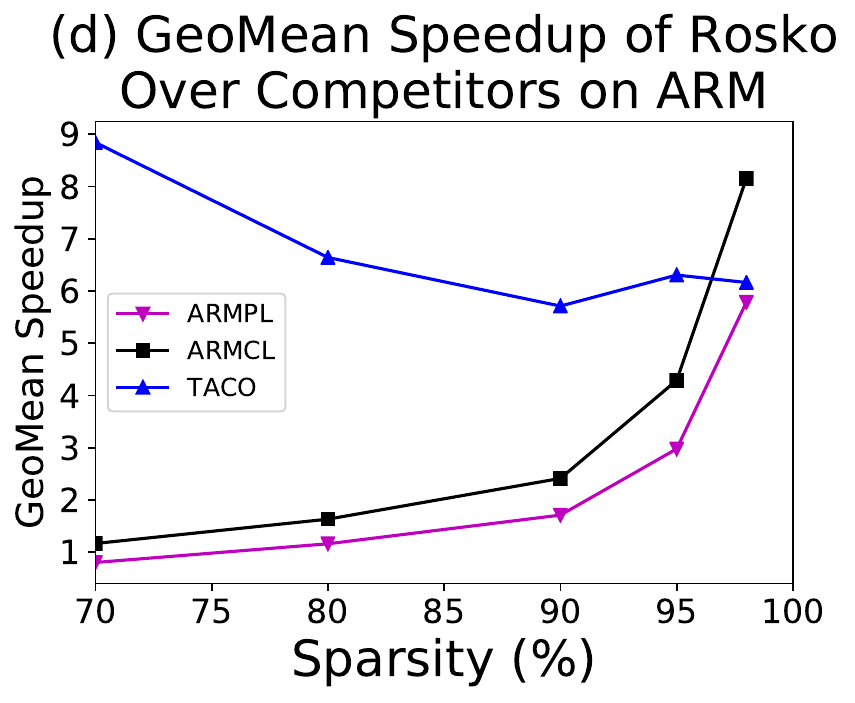}}}\hfill

\caption{
We compare Rosko kernels to competing dense and sparse kernels on Intel and ARM CPUs for SpMM  in the DLMC benchmark \cite{dlmc}.
The benchmark contains pruned weight matrices from the transformer model with sparsities from 70\% to 98\%.
In (a) and (b), each point represents an individual SpMM problem (e.g., in multi-head attention or MLP blocks) with a certain number of nonzeros.
Figures (c) and (d) summarize the geometric mean speedup of Rosko over competing methods as sparsity changes from 70\% to 98\%.
Rosko outperforms MKL's dense MM when sparsity exceeds $\approx 75\%$.
Performance of MKL-CSR SpMM improves at higher sparsities since MKL's SpMM is ideal for matrices with sparsities $>99\%$
} \label{fig:rosko_comparisons}
    \vspace{-4mm}
\end{figure*}
\begin{figure*}[ht]
    \subfloat{\label{fig:arm_io}{\includegraphics[width=0.3\textwidth]{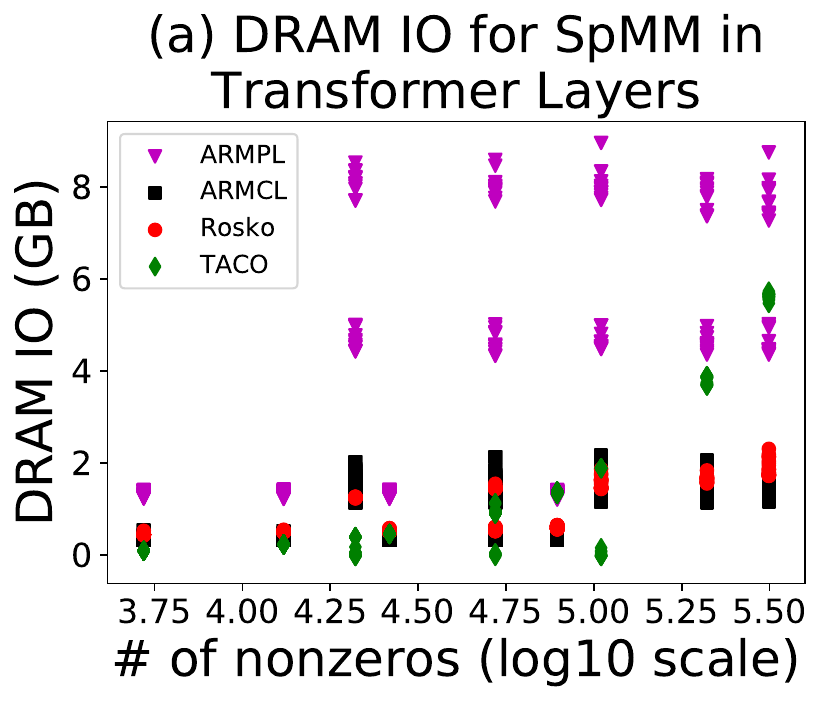}}}\hfill
    \subfloat{\label{fig:arm_bw}{\includegraphics[width=0.32\textwidth]{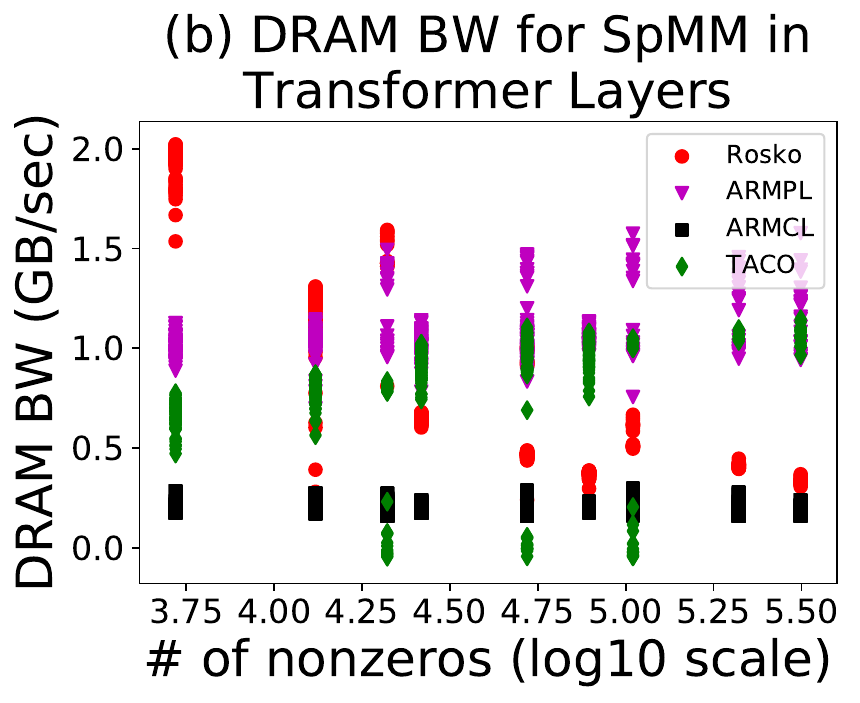}}}\hfill
    \subfloat{\label{fig:arm_dram_tput}{\includegraphics[width=0.32\textwidth]{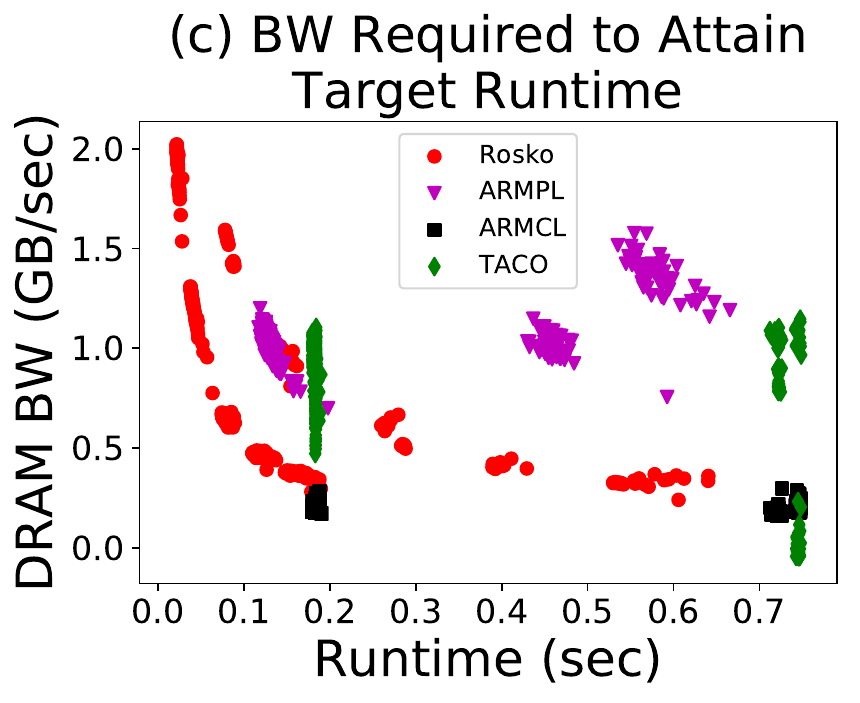}}}\hfill

\caption{
Off-chip (DRAM) memory accesses incurred during SpMM in the DLMC benchmark on the ARM Cortex-A53 CPU.
Rosko performs minimal DRAM IO accross all sparsities (a) while utilizing more DRAM BW at higher sparsities and vice versa for lower sparsities (b).
Hence, Rosko attains a better tradeoff of higher DRAM bandwidth usage for lower runtime overhead (c).
} \label{fig:arm_google_bench}
    \vspace{-4mm}
\end{figure*}

We iteratively prune a given dense model to a final sparsity of 90-97\% and evaluate performance based on runtime. Every 10 epochs, weights with the smallest magnitude are pruned until the target sparsity is attained. 
For the plots in \Cref{fig:arm_training} we pruned the smallest 10\% of nonzero weights every 10 epochs until the desired sparsity was achieved. 
We then continue training the network until a total of 200 epochs have elapsed (including the pruning epochs). 
Rosko is used for SpMM during training once weight matrices reach a 75\% sparsity level while ARMPL is used for dense MM at lower sparsities.
In these experiments, we include overhead incurred by transposes, im2col transformations, and re-packing sparse weight matrices during pruning epochs.
Both dense and sparse networks attain target accuracies of 93\% for VGG, 92.5\% for MobileNetV2, and 95.2\% for ResNet-18 after training.

Rosko achieves an $\approx 1.5$x speedup over ARMPL for training on all networks. 
We begin to use Rosko SpMM operations once sparsity reaches 70\% in epoch 70.  
Since Rosko accelerates roughly half the total epochs (130/200), we achieve roughly half of the theoretical peak speedup, as expected.

\subsection{Performance on Graph Neural Networks}
\label{sec:featgraph}

We compare Rosko to FeatGraph \cite{featgraph} and MKL-CSR \cite{intelmkl} in terms of computation throughput when performing SpMM during the aggregation phase of graph convolutional networks (GCN).
FeatGraph is a programming framework for GCNs that provides efficient SpMM kernels implemented in the TVM IR \cite{tvm}.
We measure SpMM throughput on the Intel i9 CPU using two graph datasets, Reddit \cite{reddit} and Ogbn-proteins \cite{ogb} with sparsities of 99.79 and 99.55, respectively.
For each new graph problem, FeatGraph must perform a grid search accross all possible combinations of 2 parameters, namely the graph and feature partitioning factors, to identify the best-performing combination.
In contrast, Rosko analytically derives optimal sparse tiling parameters without searching or applying graph-specific partitioning optimizations.

\begin{figure}[h]
    \centering
    \subfloat{\label{fig:reddit}{\includegraphics[scale=0.37]{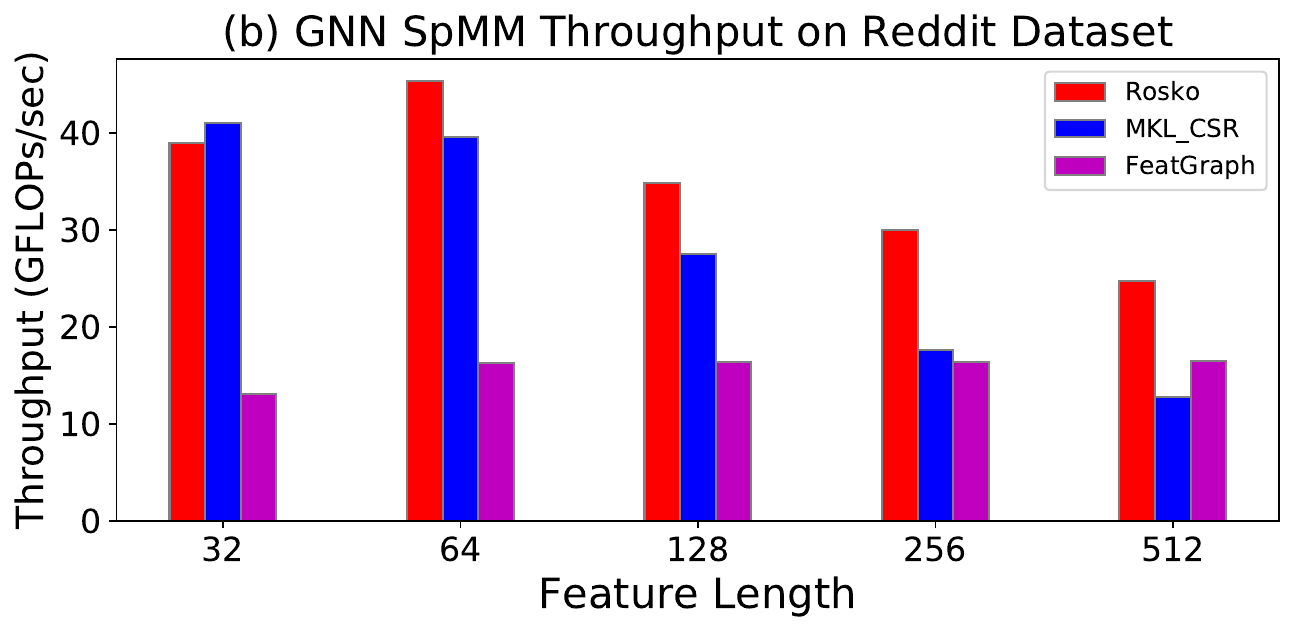}}}\\
    \subfloat{\label{fig:ogbn}{\includegraphics[scale=0.37]{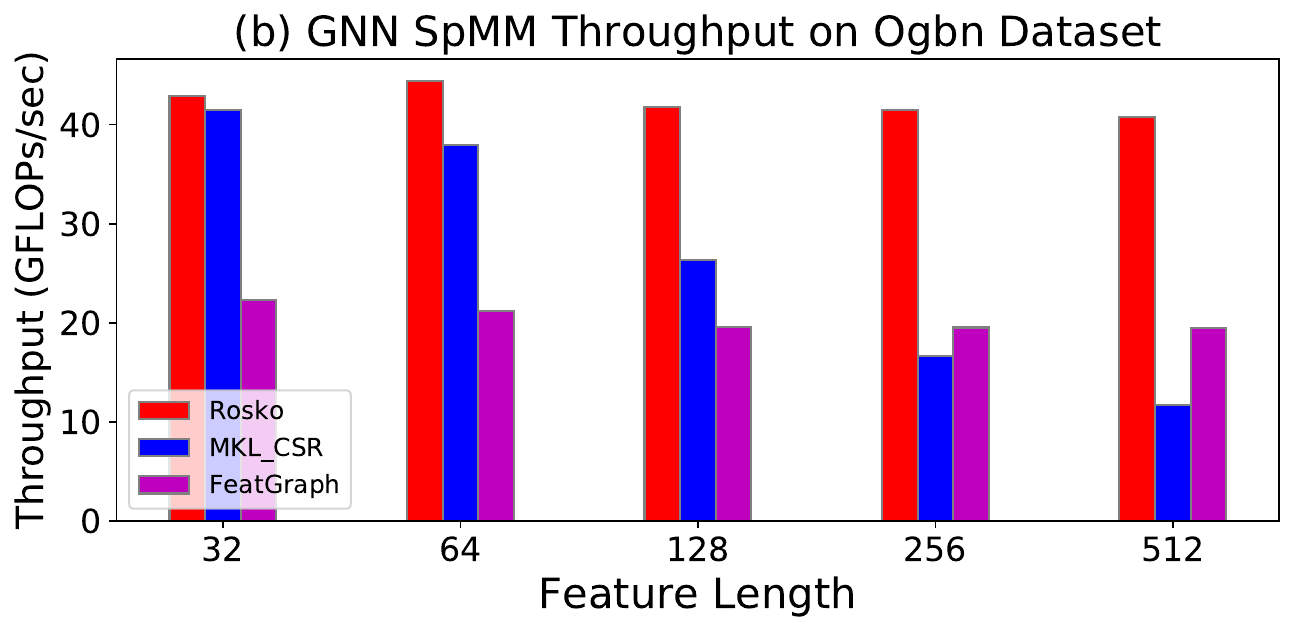}}}
\caption{
SpMM throughput during GCN aggregation using the Reddit and Ogbn-proteins datasets on all 10 cores of the Intel i9 10900K CPU.
Rosko outperforms both FeatGraph and MKL-CSR accross feature lengths.
}
\label{fig:GCN}
\end{figure}

\begin{figure}[h]
    \centering
    \subfloat{\label{fig:ogbn_tput_vs_cores}{\includegraphics[scale=0.37]{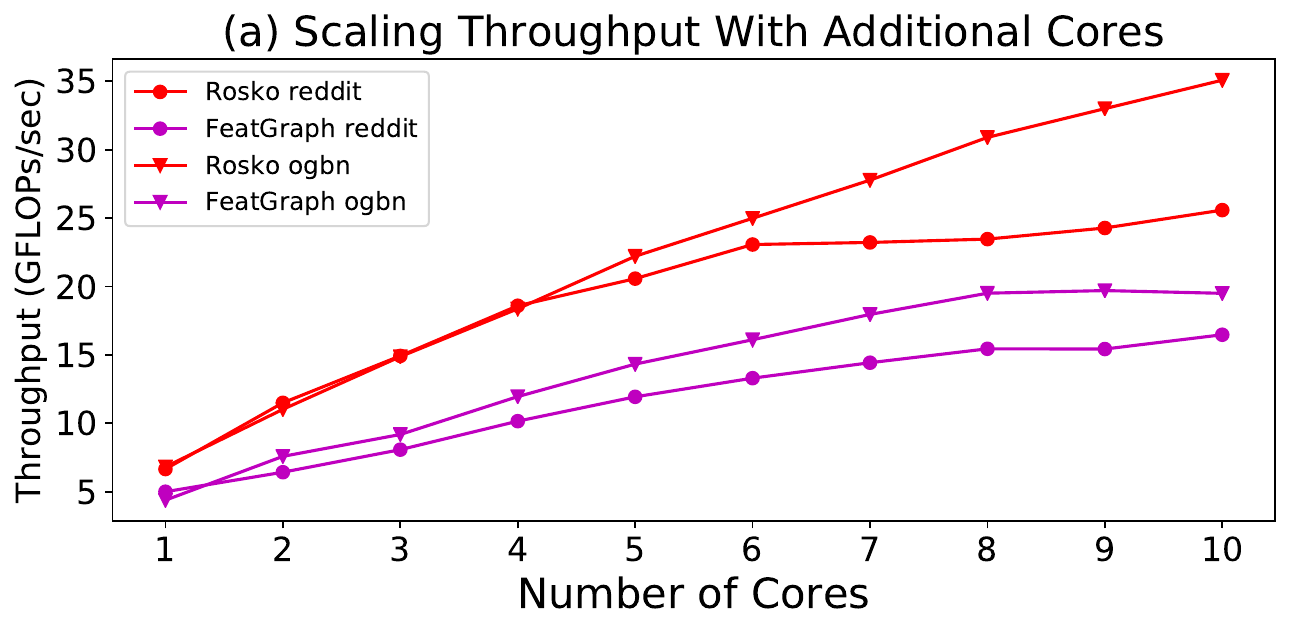}}}\\
    \subfloat{\label{fig:ogbn_throughput_vs_cores}{\includegraphics[scale=0.37]{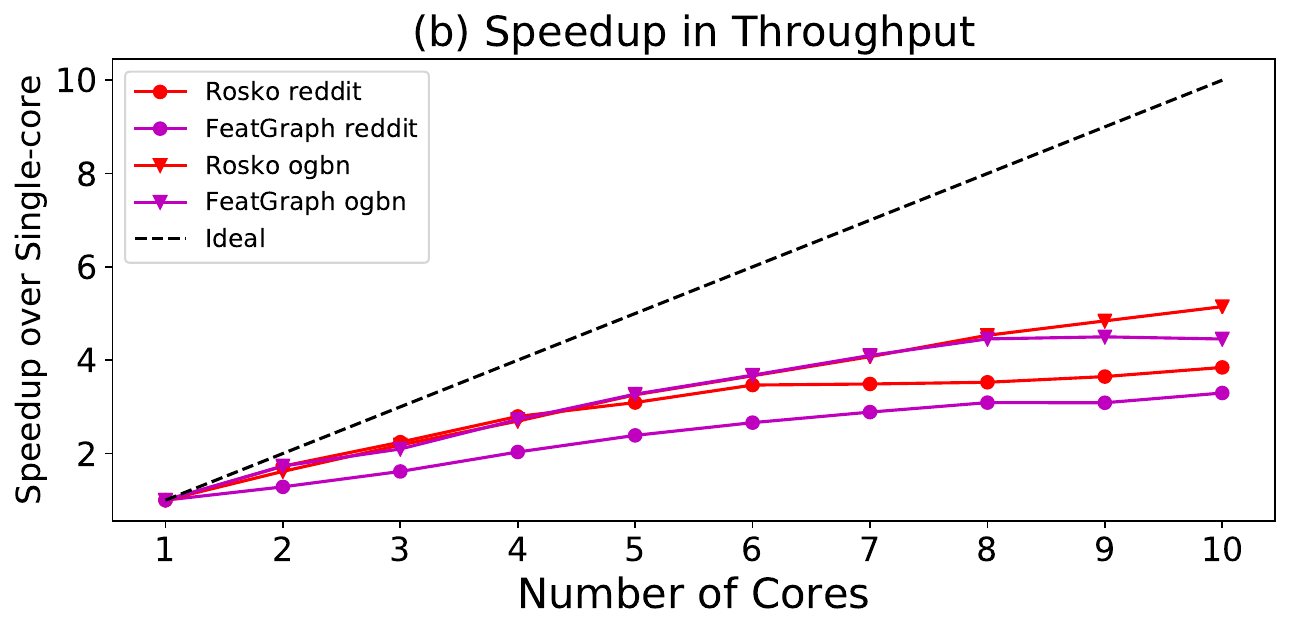}}}
\caption{ 
Rosko achieves higher throughput at all core counts (a) and better scalability (b) than FeatGraph. 
}
\label{fig:feat_load}
\end{figure}

Figures \ref{fig:reddit} and \ref{fig:ogbn} plot SpMM throughput during GCN aggregation for feature lengths ranging from 32 to 512. 
Rosko outperforms FeatGraph by 1.5x - 3.0x and MKL by 0.95x - 3.5x on both datasets when using all 10 cores on the Intel i9 CPU. 
In \Cref{fig:feat_load}, we assess the scalability of Rosko and Featgraph when fixing feature length at 512 and growing the number of cores from 1 to 10.  
In contrast to FeatGraph, Rosko shapes sparse graph tiles directly according to available system resources, achieving higher computation throughput and better scaling across all core counts. 

\section{Discussion}
\label{sec:discussion}
In this section, we discuss potential applications of Rosko as well as extensions and improvements over its current design.

\textbf{Lottery Ticket Searching}: Recent research has attempted to derive sparse sub-networks which achieve high model accuracy during early epochs of training, dubbed{
``lottery tickets'' \cite{frankle2019lottery, chen2021earlybert, you2022drawing, evci2020rigging}.}
Rosko could help improve performance of searching for lottery tickets, which involves several rounds of training weight matrices with unstructured sparsity.

\textbf{Extensions to Rosko}: Rosko's tiling model and uniform random sparsity assumptions are sufficient to achieve high performance, despite not being targeted for sparsity patterns in GNNs.
However, we note that Rosko's scaling in \Cref{fig:feat_load} may be improved by integrating with prior inspector-executor methods \cite{polyinspect,inspect}. 
Such approaches analyze sparsity patterns in the input graph before applying transformations such as row-reordering.
For instance, reordering rows may improve load balance across cores by evenly distributing nonzeros throughout the sparse matrix such that each Rosko tile has similar density.  

While Rosko achieves high performance on the selected neural network benchmarks without auto-tuning, we expose the tiling parameters $m_c$, $k_c$, $m_r$, and $n_r$ to the programmer.
This enables future works to experiment with auto-tuning our kernels for potential performance benefits beyond Rosko on other benchmarks with different sparsity patterns.

\textbf{Rosko on Other Hardware}: We anticipate porting 32-bit floating point (FP32) Rosko kernels to GPUs to be straightforward.
For dense FP32 MM on GPUs, state-of-the-art libraries like cuBLAS \cite{cublas} use outer-product-based GEMM and could benefit from row skipping on sparse inputs.
Existing sparse accelerators are inner-product-based (e.g., sparse tensor cores) and recent works have identified a need for outer product hardware, e.g., outer product tensor cores for dense MM \cite{wang2021dualside}).
In addition, several recent CPUs include custom outer product hardware to accelerate GEMM-like operations \cite{armsme, apple}.
Rosko kernels could leverage such dense outer product hardware.




\section{Conclusion}
\label{sec:conclusion}
This paper presents the Rosko row skipping algorithm that leverages outer product column sparsity.
For each of the outer products that form a matrix multiplication, we skip computation for rows corresponding to zero entries in the input column.
Rosko is simple and efficient in sparsity management: index arrays containing locations of nonzero values enable the skipping of entire rows of operations, which correspond to zero entries.
Rosko can naturally leverage SIMD architectures and multiple cores with high data reuse throughout the memory hierarchy.
Rosko can work in tandem with efficient schedules, such as block shaping, which further promote data reuse and reduce IO requirements.

With all these attributes, Rosko kernels can outperform other state-of-the-art dense and sparse matrix multiplication libraries over wide ranges of sparsities, as empirically demonstrated.
In conclusion, Rosko has provided new insights and a novel approach to the significant problem of managing the computation of sparse neural networks.

\section{Appendix}

\textbf{Rosko: Row Skipping Outer Products for Sparse
Matrix Multiplication Kernels}\\
Vikas Natesh, Andrew Sabot, H.T. Kung, Mark Ting\\
Harvard University\\

We provide details on the computational artifacts for the submission "Rosko: Row Skipping Outer Products for Sparse
Matrix Multiplication Kernels". Rosko tiles sparse-times-dense matrix multiplcation (SpMM) computations according to available system resources such as DRAM bandwidth, on-chip memory, and processing cores to attain high throughput on CPUs. Our kernels are targeted to neural network applications where matrix sparsity ranges from 65\% to 99.5\%. We implemented Rosko in C++ and SIMD intrinsics and use this implementation to compare with competing methods for dense MM and SpMM.
The Rosko SpMM library and experiments can be found at \url{https://github.com/vnatesh/Rosko.git}.
The github repo also includes code to install software of competing methods and download various benchmark datasets.

\subsection{Platform Details}
Below is a list of CPU platforms used in experiments along with their OS version, compilers, profiling tools, and external libraries, respectively:

\begin{itemize}
   \item Alienware Aurora r11 Desktop with Intel i9 10900K CPU
   \begin{itemize}
     \item Ubuntu 20.04.1 running Linux
kernel 5.8.0-48-generic  
     \item gcc-9.3.0, OpenMP 4.5
     \item Vtune 2021.1.1
     \item MKL 2021.1.1, TACO (latest github version), FeatGraph (latest github version)
    \end{itemize}

   \item Raspberry Pi 3 Model B with an ARM
v8 Cortex A53 CPU

   \begin{itemize}
     \item Ubuntu 21.04.2 running Linux
Kernel 5.11.0-1027-raspi  
     \item gcc-9.3.0, OpenMP 4.5
     \item Linux perf 5.4.86
     \item ARM Performance Libraries 21.0.0 (ARMPL), ARM Compute Library 22.11 (ARMCL), TACO (latest github version)
    \end{itemize}
\end{itemize}

\subsection{Datasets}
We select matrices from the Deep Learning Matrix Collection (DLMC) benchmark, which contains several sparse weight matrices from Transformer model layers. 
We specifically profile SpMM using DLMC matrices with sparsities from 70\% to 98\%.
For graph convolutional network (GCN) experiments, we use the Reddit and Ogbn-proteins datasets.
CNN experiments performed training over the CIFAR10 dataset.
In several experiments, we also use synthetic sparse matrices with varying levels of random uniform sparsity.
All matrices across datasets contain single-precision floating point values.

\subsection{Experiment Details}
We compare the performance of Rosko kernels to MKL dense MM, MKL-CSR SpMM, ARMPL, ARMCL, FeatGraph, and TACO on the above platforms.
Performance is measured in terms of computation throughput (GFLOPs/sec), where the number of FLOPs is computed as $d \cdot MKN = \#nnz \cdot N$, where $d$ is the fraction of values that are nonzero in the $M\times K$ sparse matrix. 
For memory usage, we report DRAM bandwidth (GB/sec) and total DRAM IO (GB) as our metrics. 
Each figure in the submission has a corresponding experiment located in the relevant directory (e.g., experiments/intel/packing contains the packing experiment of Fig. 6).
Each experiment directory contains 3 files:
   \begin{itemize}
     \item install.sh for installing competing method software and dependencies
     \item run.sh for downloading datasets and running the experiment to generate results (csv and/or text files)
     \item plots.py for generating the corresponding PDF figure from results
    \end{itemize}

Below is the comprehensive list of figures and corresponding experiments.
   \begin{itemize}
     \item Figure 6 compares single-core packing performance of Rosko and MKL-CSR on synthetic sparse matrices
     \item Figure 8 validates DRAM bandwidth usage and throughput of Rosko on both platforms when performing SpMM with synthetic sparse matrices on multiple cores
     \item Figure 9 shows Rosko's speedup in runtime over ARMPL when training 3 different CNNs on the CIFAR10 dataset on the ARM CPU.
     \item In Figure 10, we compare Rosko's throughput to MKL-dense, MKL-CSR, and TACO on the Intel CPU and ARMPL, ARMCL, and TACO on the ARM CPU when performing SpMM in the DLMC benchmark
     \item Figure 11 plots DRAM bandwidth usage, on the ARM CPU, of Rosko, ARMPL, ARMCL, and TACO during SpMM in the DLMC benchmark.
     \item Figures 12 and 13 plot throughput, on the Intel CPU, of Rosko, MKL-CSR, and FeatGraph when running SpMM during GCN aggregation.
    \end{itemize}

In all experiments, we average our performance measurements over 50 trials.
We flush the cache between runs to avoid cache reuse across successive runs and measure the execution time of each individual run.
We also disable simultaneous multithreading (SMT) and dynamic voltage and frequency scaling (DVFS) on each system to maintain a consistent clock frequency and reduce variability across runs. 
Due the size of the benchmark datasets as well as overhead of cache flushing between every run, experiments may take 4-16 hours to run, depending on platform.
For testing purposes, experiment runtime may be reduced by reducing the number of trials or disabling cache flushing in the run.sh file, although this may impact accuracy of the results.


\bibliographystyle{ACM-Reference-Format}
\bibliography{references, refs}


\begin{thebibliography}{50}


\ifx \showCODEN    \undefined \def \showCODEN     #1{\unskip}     \fi
\ifx \showDOI      \undefined \def \showDOI       #1{#1}\fi
\ifx \showISBNx    \undefined \def \showISBNx     #1{\unskip}     \fi
\ifx \showISBNxiii \undefined \def \showISBNxiii  #1{\unskip}     \fi
\ifx \showISSN     \undefined \def \showISSN      #1{\unskip}     \fi
\ifx \showLCCN     \undefined \def \showLCCN      #1{\unskip}     \fi
\ifx \shownote     \undefined \def \shownote      #1{#1}          \fi
\ifx \showarticletitle \undefined \def \showarticletitle #1{#1}   \fi
\ifx \showURL      \undefined \def \showURL       {\relax}        \fi
\providecommand\bibfield[2]{#2}
\providecommand\bibinfo[2]{#2}
\providecommand\natexlab[1]{#1}
\providecommand\showeprint[2][]{arXiv:#2}

\bibitem[arm(2021)]%
        {armsme}
 \bibinfo{year}{2021}\natexlab{}.
\newblock \bibinfo{title}{{{ARM SME}}}.
\newblock
  \bibinfo{howpublished}{https://community.arm.com/arm-community-blogs/b/architectures-and-processors-blog/posts/scalable-matrix-extension-armv9-a-architecture}.
\newblock


\bibitem[per(2021)]%
        {perf}
 \bibinfo{year}{2021}\natexlab{}.
\newblock \bibinfo{title}{perf: Linux profiling with performance counters}.
\newblock
  \bibinfo{howpublished}{\url{https://perf.wiki.kernel.org/index.php/Main_Page}}.
\newblock


\bibitem[int(2023)]%
        {intelmkl}
 \bibinfo{year}{2023}\natexlab{}.
\newblock \bibinfo{title}{Intel oneAPI Math Kernel Library}.
\newblock
  \bibinfo{howpublished}{\url{https://software.intel.com/content/www/us/en/develop/articles/mkl-reference-manual.html}}.
\newblock


\bibitem[Ahrens et~al\mbox{.}(2022)]%
        {tacosched}
\bibfield{author}{\bibinfo{person}{Peter Ahrens}, \bibinfo{person}{Fredrik
  Kjolstad}, {and} \bibinfo{person}{Saman Amarasinghe}.}
  \bibinfo{year}{2022}\natexlab{}.
\newblock \showarticletitle{Autoscheduling for Sparse Tensor Algebra with an
  Asymptotic Cost Model}. In \bibinfo{booktitle}{\emph{Proceedings of the 43rd
  ACM SIGPLAN International Conference on Programming Language Design and
  Implementation}} (San Diego, CA, USA) \emph{(\bibinfo{series}{PLDI 2022})}.
  \bibinfo{publisher}{Association for Computing Machinery},
  \bibinfo{address}{New York, NY, USA}, \bibinfo{pages}{269–285}.
\newblock
\showISBNx{9781450392655}
\urldef\tempurl%
\url{https://doi.org/10.1145/3519939.3523442}
\showDOI{\tempurl}


\bibitem[Anwar et~al\mbox{.}(2017)]%
        {anwar2017structured}
\bibfield{author}{\bibinfo{person}{Sajid Anwar}, \bibinfo{person}{Kyuyeon
  Hwang}, {and} \bibinfo{person}{Wonyong Sung}.}
  \bibinfo{year}{2017}\natexlab{}.
\newblock \showarticletitle{Structured {{Pruning}} of {{Deep Convolutional
  Neural Networks}}}.
\newblock \bibinfo{journal}{\emph{ACM Journal on Emerging Technologies in
  Computing Systems}} \bibinfo{volume}{13}, \bibinfo{number}{3}
  (\bibinfo{date}{Feb.} \bibinfo{year}{2017}), \bibinfo{pages}{32:1--32:18}.
\newblock
\showISSN{1550-4832}
\urldef\tempurl%
\url{https://doi.org/10.1145/3005348}
\showDOI{\tempurl}


\bibitem[{Apple Inc.}(2022)]%
        {apple}
\bibfield{author}{\bibinfo{person}{{Apple Inc.}}}
  \bibinfo{year}{2022}\natexlab{}.
\newblock \bibinfo{title}{Deploying Transformers on the Apple Neural Engine}.
\newblock
  \bibinfo{howpublished}{https://machinelearning.apple.com/research/neural-engine-transformers}.
\newblock


\bibitem[Arm Limited(2021)]%
        {armpl}
Arm Limited \bibinfo{year}{2021}\natexlab{}.
\newblock \bibinfo{booktitle}{\emph{Arm Performance Libraries Reference
  Guide}}.
\newblock Arm Limited.
\newblock
\urldef\tempurl%
\url{https://developer.arm.com/documentation/101004/latest/}
\showURL{%
\tempurl}


\bibitem[Arm Limited(2022)]%
        {armcl}
Arm Limited \bibinfo{year}{2022}\natexlab{}.
\newblock \bibinfo{booktitle}{\emph{Arm Compute Library Reference Guide}}.
\newblock Arm Limited.
\newblock
\urldef\tempurl%
\url{https://arm-software.github.io/ComputeLibrary/latest/}
\showURL{%
\tempurl}


\bibitem[Chen et~al\mbox{.}(2018)]%
        {tvm}
\bibfield{author}{\bibinfo{person}{Tianqi Chen}, \bibinfo{person}{Thierry
  Moreau}, \bibinfo{person}{Ziheng Jiang}, \bibinfo{person}{Lianmin Zheng},
  \bibinfo{person}{Eddie~Q. Yan}, \bibinfo{person}{Haichen Shen},
  \bibinfo{person}{Meghan Cowan}, \bibinfo{person}{Leyuan Wang},
  \bibinfo{person}{Yuwei Hu}, \bibinfo{person}{Luis Ceze},
  \bibinfo{person}{Carlos Guestrin}, {and} \bibinfo{person}{Arvind
  Krishnamurthy}.} \bibinfo{year}{2018}\natexlab{}.
\newblock \showarticletitle{TVM: An Automated End-to-End Optimizing Compiler
  for Deep Learning}. In \bibinfo{booktitle}{\emph{USENIX Symposium on
  Operating Systems Design and Implementation}}.
\newblock


\bibitem[Chen et~al\mbox{.}(2021)]%
        {chen2021earlybert}
\bibfield{author}{\bibinfo{person}{Xiaohan Chen}, \bibinfo{person}{Yu Cheng},
  \bibinfo{person}{Shuohang Wang}, \bibinfo{person}{Zhe Gan},
  \bibinfo{person}{Zhangyang Wang}, {and} \bibinfo{person}{Jingjing Liu}.}
  \bibinfo{year}{2021}\natexlab{}.
\newblock \showarticletitle{{{EarlyBERT}}: {{Efficient BERT Training}} via
  {{Early-bird Lottery Tickets}}}. In \bibinfo{booktitle}{\emph{Proceedings of
  the 59th {{Annual Meeting}} of the {{Association}} for {{Computational
  Linguistics}} and the 11th {{International Joint Conference}} on {{Natural
  Language Processing}} ({{Volume}} 1: {{Long Papers}})}}.
  \bibinfo{publisher}{{Association for Computational Linguistics}},
  \bibinfo{address}{{Online}}, \bibinfo{pages}{2195--2207}.
\newblock
\urldef\tempurl%
\url{https://doi.org/10.18653/v1/2021.acl-long.171}
\showDOI{\tempurl}


\bibitem[Cheshmi et~al\mbox{.}(2022)]%
        {inspect}
\bibfield{author}{\bibinfo{person}{Kazem Cheshmi}, \bibinfo{person}{Zachary
  Cetinic}, {and} \bibinfo{person}{Maryam~Mehri Dehnavi}.}
  \bibinfo{year}{2022}\natexlab{}.
\newblock \showarticletitle{Vectorizing Sparse Matrix Computations with
  Partially-Strided Codelets}. In \bibinfo{booktitle}{\emph{Proceedings of the
  International Conference on High Performance Computing, Networking, Storage
  and Analysis}} (Dallas, Texas) \emph{(\bibinfo{series}{SC '22})}.
  \bibinfo{publisher}{IEEE Press}, Article \bibinfo{articleno}{32},
  \bibinfo{numpages}{15}~pages.
\newblock
\showISBNx{9784665454445}


\bibitem[Corporation(2021)]%
        {VTune}
\bibfield{author}{\bibinfo{person}{Intel Corporation}.}
  \bibinfo{year}{2021}\natexlab{}.
\newblock \bibinfo{title}{Intel VTune Profiler}.
\newblock
  \bibinfo{howpublished}{\url{https://software.intel.com/content/www/us/en/develop/tools/oneapi/components/vtune-profiler.html}}.
\newblock


\bibitem[Corporation(2022)]%
        {cublas}
\bibfield{author}{\bibinfo{person}{NVIDIA Corporation}.}
  \bibinfo{year}{2022}\natexlab{}.
\newblock \bibinfo{title}{cuBLAS}.
\newblock \bibinfo{howpublished}{\url{https://developer.nvidia.com/cublas}}.
\newblock


\bibitem[Dongarra(1995a)]%
        {dongarraccs}
\bibfield{author}{\bibinfo{person}{Jack Dongarra}.}
  \bibinfo{year}{1995}\natexlab{a}.
\newblock \bibinfo{title}{Compressed {{Column Storage}} ({{CCS}})}.
\newblock
  \bibinfo{howpublished}{\url{http://netlib.org/linalg/html\_templates/node92.html}}.
\newblock


\bibitem[Dongarra(1995b)]%
        {dongarracrs}
\bibfield{author}{\bibinfo{person}{Jack Dongarra}.}
  \bibinfo{year}{1995}\natexlab{b}.
\newblock \bibinfo{title}{Compressed Row Storage (CRS)}.
\newblock
  \bibinfo{howpublished}{\url{http://netlib.org/linalg/html\_templates/node91.html}}.
\newblock


\bibitem[Evci et~al\mbox{.}(2020)]%
        {evci2020rigging}
\bibfield{author}{\bibinfo{person}{Utku Evci}, \bibinfo{person}{Trevor Gale},
  \bibinfo{person}{Jacob Menick}, \bibinfo{person}{Pablo~Samuel Castro}, {and}
  \bibinfo{person}{Erich Elsen}.} \bibinfo{year}{2020}\natexlab{}.
\newblock \showarticletitle{Rigging the Lottery: Making All Tickets Winners}.
  In \bibinfo{booktitle}{\emph{Proceedings of the 37th {{International
  Conference}} on {{Machine Learning}}}}
  \emph{(\bibinfo{series}{{{ICML}}'20})}. \bibinfo{publisher}{{JMLR.org}},
  \bibinfo{pages}{2943--2952}.
\newblock


\bibitem[Frankle and Carbin(2019)]%
        {frankle2019lottery}
\bibfield{author}{\bibinfo{person}{Jonathan Frankle} {and}
  \bibinfo{person}{Michael Carbin}.} \bibinfo{year}{2019}\natexlab{}.
\newblock \showarticletitle{The {{Lottery Ticket Hypothesis}}: {{Finding
  Sparse}}, {{Trainable Neural Networks}}}.
\newblock \bibinfo{journal}{\emph{arXiv:1803.03635 [cs]}}
  (\bibinfo{date}{March} \bibinfo{year}{2019}).
\newblock
\showeprint[arxiv]{1803.03635}~[cs]


\bibitem[Gale et~al\mbox{.}(2020)]%
        {gale2020sparse}
\bibfield{author}{\bibinfo{person}{Trevor Gale}, \bibinfo{person}{Matei
  Zaharia}, \bibinfo{person}{Cliff Young}, {and} \bibinfo{person}{Erich
  Elsen}.} \bibinfo{year}{2020}\natexlab{}.
\newblock \showarticletitle{Sparse {{GPU}} Kernels for Deep Learning}. In
  \bibinfo{booktitle}{\emph{Proceedings of the {{International Conference}} for
  {{High Performance Computing}}, {{Networking}}, {{Storage}} and
  {{Analysis}}}} \emph{(\bibinfo{series}{{{SC}} '20})}.
  \bibinfo{publisher}{{IEEE Press}}, \bibinfo{address}{{Atlanta, Georgia}},
  \bibinfo{pages}{1--14}.
\newblock
\showISBNx{978-1-72819-998-6}


\bibitem[Goto and Geijn(2008)]%
        {gotomain}
\bibfield{author}{\bibinfo{person}{Kazushige Goto} {and}
  \bibinfo{person}{Robert A. van~de Geijn}.} \bibinfo{year}{2008}\natexlab{}.
\newblock \showarticletitle{Anatomy of High-Performance Matrix Multiplication}.
\newblock \bibinfo{journal}{\emph{ACM Trans. Math. Softw.}}, Article
  \bibinfo{articleno}{12} (\bibinfo{year}{2008}), \bibinfo{numpages}{25}~pages.
\newblock


\bibitem[Hamilton et~al\mbox{.}(2017)]%
        {reddit}
\bibfield{author}{\bibinfo{person}{William~L. Hamilton}, \bibinfo{person}{Rex
  Ying}, {and} \bibinfo{person}{Jure Leskovec}.}
  \bibinfo{year}{2017}\natexlab{}.
\newblock \showarticletitle{Inductive Representation Learning on Large Graphs}.
  In \bibinfo{booktitle}{\emph{Proceedings of the 31st International Conference
  on Neural Information Processing Systems}} (Long Beach, California, USA)
  \emph{(\bibinfo{series}{NIPS'17})}. \bibinfo{publisher}{Curran Associates
  Inc.}, \bibinfo{address}{Red Hook, NY, USA}, \bibinfo{pages}{1025–1035}.
\newblock
\showISBNx{9781510860964}


\bibitem[Han et~al\mbox{.}(2015)]%
        {han2015compression}
\bibfield{author}{\bibinfo{person}{Song Han}, \bibinfo{person}{Huizi Mao},
  {and} \bibinfo{person}{William~J. Dally}.} \bibinfo{year}{2015}\natexlab{}.
\newblock \showarticletitle{Deep Compression: Compressing Deep Neural Networks
  with Pruning, Trained Quantization and Huffman Coding}.
\newblock  (\bibinfo{year}{2015}).
\newblock
\urldef\tempurl%
\url{http://arxiv.org/abs/1510.00149}
\showURL{%
\tempurl}
\newblock
\shownote{cite arxiv:1510.00149Comment: Published as a conference paper at ICLR
  2016 (oral)}.


\bibitem[He et~al\mbox{.}(2015a)]%
        {he2015deep}
\bibfield{author}{\bibinfo{person}{Kaiming He}, \bibinfo{person}{Xiangyu
  Zhang}, \bibinfo{person}{Shaoqing Ren}, {and} \bibinfo{person}{Jian Sun}.}
  \bibinfo{year}{2015}\natexlab{a}.
\newblock \bibinfo{title}{Deep Residual Learning for Image Recognition}.
\newblock
\newblock
\showeprint[arxiv]{1512.03385}~[cs.CV]


\bibitem[He et~al\mbox{.}(2015b)]%
        {resnet}
\bibfield{author}{\bibinfo{person}{Kaiming He}, \bibinfo{person}{Xiangyu
  Zhang}, \bibinfo{person}{Shaoqing Ren}, {and} \bibinfo{person}{Jian Sun}.}
  \bibinfo{year}{2015}\natexlab{b}.
\newblock \bibinfo{title}{Deep Residual Learning for Image Recognition}.
\newblock
\newblock
\urldef\tempurl%
\url{https://doi.org/10.48550/ARXIV.1512.03385}
\showDOI{\tempurl}


\bibitem[Howard et~al\mbox{.}(2017)]%
        {howard2017mobilenets}
\bibfield{author}{\bibinfo{person}{Andrew~G. Howard}, \bibinfo{person}{Menglong
  Zhu}, \bibinfo{person}{Bo Chen}, \bibinfo{person}{Dmitry Kalenichenko},
  \bibinfo{person}{Weijun Wang}, \bibinfo{person}{Tobias Weyand},
  \bibinfo{person}{Marco Andreetto}, {and} \bibinfo{person}{Hartwig Adam}.}
  \bibinfo{year}{2017}\natexlab{}.
\newblock \bibinfo{title}{MobileNets: Efficient Convolutional Neural Networks
  for Mobile Vision Applications}.
\newblock
\newblock
\showeprint[arxiv]{1704.04861}~[cs.CV]


\bibitem[Hu et~al\mbox{.}(2020a)]%
        {ogb}
\bibfield{author}{\bibinfo{person}{Weihua Hu}, \bibinfo{person}{Matthias Fey},
  \bibinfo{person}{Marinka Zitnik}, \bibinfo{person}{Yuxiao Dong},
  \bibinfo{person}{Hongyu Ren}, \bibinfo{person}{Bowen Liu},
  \bibinfo{person}{Michele Catasta}, {and} \bibinfo{person}{Jure Leskovec}.}
  \bibinfo{year}{2020}\natexlab{a}.
\newblock \showarticletitle{Open Graph Benchmark: Datasets for Machine Learning
  on Graphs}. In \bibinfo{booktitle}{\emph{Advances in Neural Information
  Processing Systems}}, \bibfield{editor}{\bibinfo{person}{H.~Larochelle},
  \bibinfo{person}{M.~Ranzato}, \bibinfo{person}{R.~Hadsell},
  \bibinfo{person}{M.F. Balcan}, {and} \bibinfo{person}{H.~Lin}} (Eds.),
  Vol.~\bibinfo{volume}{33}. \bibinfo{publisher}{Curran Associates, Inc.},
  \bibinfo{pages}{22118--22133}.
\newblock
\urldef\tempurl%
\url{https://proceedings.neurips.cc/paper_files/paper/2020/file/fb60d411a5c5b72b2e7d3527cfc84fd0-Paper.pdf}
\showURL{%
\tempurl}


\bibitem[Hu et~al\mbox{.}(2020b)]%
        {featgraph}
\bibfield{author}{\bibinfo{person}{Yuwei Hu}, \bibinfo{person}{Zihao Ye},
  \bibinfo{person}{Minjie Wang}, \bibinfo{person}{Jiali Yu},
  \bibinfo{person}{Da Zheng}, \bibinfo{person}{Mu Li}, \bibinfo{person}{Zheng
  Zhang}, \bibinfo{person}{Zhiru Zhang}, {and} \bibinfo{person}{Yida Wang}.}
  \bibinfo{year}{2020}\natexlab{b}.
\newblock \showarticletitle{FeatGraph: A Flexible and Efficient Backend for
  Graph Neural Network Systems}. In \bibinfo{booktitle}{\emph{Proceedings of
  the International Conference for High Performance Computing, Networking,
  Storage and Analysis}} (Atlanta, Georgia) \emph{(\bibinfo{series}{SC '20})}.
  \bibinfo{publisher}{IEEE Press}, Article \bibinfo{articleno}{71},
  \bibinfo{numpages}{13}~pages.
\newblock
\showISBNx{9781728199986}


\bibitem[Ivanov et~al\mbox{.}(2020)]%
        {transformer_contractions}
\bibfield{author}{\bibinfo{person}{Andrei Ivanov}, \bibinfo{person}{Nikoli
  Dryden}, \bibinfo{person}{Tal Ben-Nun}, \bibinfo{person}{Shigang Li}, {and}
  \bibinfo{person}{Torsten Hoefler}.} \bibinfo{year}{2020}\natexlab{}.
\newblock \bibinfo{title}{Data Movement Is All You Need: A Case Study on
  Optimizing Transformers}.
\newblock
\newblock
\urldef\tempurl%
\url{https://doi.org/10.48550/ARXIV.2007.00072}
\showDOI{\tempurl}


\bibitem[Kjolstad et~al\mbox{.}(2017)]%
        {taco}
\bibfield{author}{\bibinfo{person}{Fredrik Kjolstad}, \bibinfo{person}{Shoaib
  Kamil}, \bibinfo{person}{Stephen Chou}, \bibinfo{person}{David Lugato}, {and}
  \bibinfo{person}{Saman Amarasinghe}.} \bibinfo{year}{2017}\natexlab{}.
\newblock \showarticletitle{The Tensor Algebra Compiler}.
\newblock \bibinfo{journal}{\emph{Proc. ACM Program. Lang.}}
  \bibinfo{volume}{1}, \bibinfo{number}{OOPSLA}, Article
  \bibinfo{articleno}{77} (\bibinfo{date}{oct} \bibinfo{year}{2017}),
  \bibinfo{numpages}{29}~pages.
\newblock
\urldef\tempurl%
\url{https://doi.org/10.1145/3133901}
\showDOI{\tempurl}


\bibitem[Koanantakool et~al\mbox{.}(2016)]%
        {bwbound}
\bibfield{author}{\bibinfo{person}{Penporn Koanantakool},
  \bibinfo{person}{Ariful Azad}, \bibinfo{person}{Aydin Buluç},
  \bibinfo{person}{Dmitriy Morozov}, \bibinfo{person}{Sang-Yun Oh},
  \bibinfo{person}{Leonid Oliker}, {and} \bibinfo{person}{Katherine Yelick}.}
  \bibinfo{year}{2016}\natexlab{}.
\newblock \showarticletitle{Communication-Avoiding Parallel Sparse-Dense
  Matrix-Matrix Multiplication}. In \bibinfo{booktitle}{\emph{2016 IEEE
  International Parallel and Distributed Processing Symposium (IPDPS)}}.
  \bibinfo{pages}{842--853}.
\newblock
\urldef\tempurl%
\url{https://doi.org/10.1109/IPDPS.2016.117}
\showDOI{\tempurl}


\bibitem[Kolodziej et~al\mbox{.}(2019)]%
        {kolodziej2019suitesparse}
\bibfield{author}{\bibinfo{person}{Scott~P Kolodziej}, \bibinfo{person}{Mohsen
  Aznaveh}, \bibinfo{person}{Matthew Bullock}, \bibinfo{person}{Jarrett David},
  \bibinfo{person}{Timothy~A Davis}, \bibinfo{person}{Matthew Henderson},
  \bibinfo{person}{Yifan Hu}, {and} \bibinfo{person}{Read Sandstrom}.}
  \bibinfo{year}{2019}\natexlab{}.
\newblock \showarticletitle{The suitesparse matrix collection website
  interface}.
\newblock \bibinfo{journal}{\emph{Journal of Open Source Software}}
  \bibinfo{volume}{4}, \bibinfo{number}{35} (\bibinfo{year}{2019}),
  \bibinfo{pages}{1244}.
\newblock


\bibitem[Krizhevsky(2009)]%
        {Krizhevsky2009LearningML}
\bibfield{author}{\bibinfo{person}{Alex Krizhevsky}.}
  \bibinfo{year}{2009}\natexlab{}.
\newblock \showarticletitle{Learning Multiple Layers of Features from Tiny
  Images}.
\newblock


\bibitem[Kung et~al\mbox{.}(2020)]%
        {termquant}
\bibfield{author}{\bibinfo{person}{H.~T. Kung}, \bibinfo{person}{Bradley
  McDanel}, {and} \bibinfo{person}{Sai~Qian Zhang}.}
  \bibinfo{year}{2020}\natexlab{}.
\newblock \showarticletitle{Term Quantization: Furthering Quantization at Run
  Time}. In \bibinfo{booktitle}{\emph{SC20: International Conference for High
  Performance Computing, Networking, Storage and Analysis}}.
  \bibinfo{pages}{1--16}.
\newblock
\urldef\tempurl%
\url{https://doi.org/10.1109/SC41405.2020.00100}
\showDOI{\tempurl}


\bibitem[Kung et~al\mbox{.}(2021)]%
        {cake}
\bibfield{author}{\bibinfo{person}{H.~T. Kung}, \bibinfo{person}{Vikas Natesh},
  {and} \bibinfo{person}{Andrew Sabot}.} \bibinfo{year}{2021}\natexlab{}.
\newblock \showarticletitle{CAKE: Matrix Multiplication Using
  Constant-Bandwidth Blocks}. In \bibinfo{booktitle}{\emph{Proceedings of the
  International Conference for High Performance Computing, Networking, Storage
  and Analysis}} (St. Louis, Missouri) \emph{(\bibinfo{series}{SC '21})}.
  \bibinfo{publisher}{Association for Computing Machinery},
  \bibinfo{address}{New York, NY, USA}, Article \bibinfo{articleno}{85},
  \bibinfo{numpages}{14}~pages.
\newblock
\showISBNx{9781450384421}
\urldef\tempurl%
\url{https://doi.org/10.1145/3458817.3476166}
\showDOI{\tempurl}


\bibitem[Liu et~al\mbox{.}(2017)]%
        {liu2017learning}
\bibfield{author}{\bibinfo{person}{Zhuang Liu}, \bibinfo{person}{Jianguo Li},
  \bibinfo{person}{Zhiqiang Shen}, \bibinfo{person}{Gao Huang},
  \bibinfo{person}{Shoumeng Yan}, {and} \bibinfo{person}{Changshui Zhang}.}
  \bibinfo{year}{2017}\natexlab{}.
\newblock \showarticletitle{Learning {{Efficient Convolutional Networks}}
  through {{Network Slimming}}}. In \bibinfo{booktitle}{\emph{2017 {{IEEE
  International Conference}} on {{Computer Vision}} ({{ICCV}})}}.
  \bibinfo{pages}{2755--2763}.
\newblock
\showISSN{2380-7504}
\urldef\tempurl%
\url{https://doi.org/10.1109/ICCV.2017.298}
\showDOI{\tempurl}


\bibitem[McDanel et~al\mbox{.}(2019)]%
        {mcdanel2019full}
\bibfield{author}{\bibinfo{person}{Bradley McDanel}, \bibinfo{person}{Sai~Qian
  Zhang}, \bibinfo{person}{HT Kung}, {and} \bibinfo{person}{Xin Dong}.}
  \bibinfo{year}{2019}\natexlab{}.
\newblock \showarticletitle{Full-stack optimization for accelerating CNNs using
  powers-of-two weights with FPGA validation}. In
  \bibinfo{booktitle}{\emph{Proceedings of the ACM International Conference on
  Supercomputing}}. \bibinfo{pages}{449--460}.
\newblock


\bibitem[Narang et~al\mbox{.}(2017)]%
        {narang2017blocksparse}
\bibfield{author}{\bibinfo{person}{Sharan Narang}, \bibinfo{person}{Eric
  Undersander}, {and} \bibinfo{person}{Gregory Diamos}.}
  \bibinfo{year}{2017}\natexlab{}.
\newblock \showarticletitle{Block-{{Sparse Recurrent Neural Networks}}}.
\newblock \bibinfo{journal}{\emph{arXiv:1711.02782 [cs, stat]}}
  (\bibinfo{date}{Nov.} \bibinfo{year}{2017}).
\newblock
\showeprint[arxiv]{1711.02782}~[cs, stat]


\bibitem[Pal et~al\mbox{.}(2018)]%
        {pal2018outerspace}
\bibfield{author}{\bibinfo{person}{Subhankar Pal}, \bibinfo{person}{Jonathan
  Beaumont}, \bibinfo{person}{Dong-Hyeon Park}, \bibinfo{person}{Aporva
  Amarnath}, \bibinfo{person}{Siying Feng}, \bibinfo{person}{Chaitali
  Chakrabarti}, \bibinfo{person}{Hun-Seok Kim}, \bibinfo{person}{David Blaauw},
  \bibinfo{person}{Trevor Mudge}, {and} \bibinfo{person}{Ronald Dreslinski}.}
  \bibinfo{year}{2018}\natexlab{}.
\newblock \showarticletitle{{{OuterSPACE}}: {{An Outer Product Based Sparse
  Matrix Multiplication Accelerator}}}. In \bibinfo{booktitle}{\emph{2018
  {{IEEE International Symposium}} on {{High Performance Computer
  Architecture}} ({{HPCA}})}}. \bibinfo{pages}{724--736}.
\newblock
\showISSN{2378-203X}
\urldef\tempurl%
\url{https://doi.org/10.1109/HPCA.2018.00067}
\showDOI{\tempurl}


\bibitem[Research(2020)]%
        {dlmc}
\bibfield{author}{\bibinfo{person}{Google Research}.}
  \bibinfo{year}{2020}\natexlab{}.
\newblock \bibinfo{title}{Deep Learning Matrix Collection}.
\newblock
  \bibinfo{howpublished}{\url{https://github.com/google-research/google-research/tree/master/sgk}}.
\newblock


\bibitem[Senanayake et~al\mbox{.}(2020)]%
        {taco_opt}
\bibfield{author}{\bibinfo{person}{Ryan Senanayake}, \bibinfo{person}{Changwan
  Hong}, \bibinfo{person}{Ziheng Wang}, \bibinfo{person}{Amalee Wilson},
  \bibinfo{person}{Stephen Chou}, \bibinfo{person}{Shoaib Kamil},
  \bibinfo{person}{Saman Amarasinghe}, {and} \bibinfo{person}{Fredrik
  Kjolstad}.} \bibinfo{year}{2020}\natexlab{}.
\newblock \showarticletitle{A Sparse Iteration Space Transformation Framework
  for Sparse Tensor Algebra}.
\newblock \bibinfo{journal}{\emph{Proc. ACM Program. Lang.}}
  \bibinfo{volume}{4}, \bibinfo{number}{OOPSLA}, Article
  \bibinfo{articleno}{158} (\bibinfo{date}{nov} \bibinfo{year}{2020}),
  \bibinfo{numpages}{30}~pages.
\newblock
\urldef\tempurl%
\url{https://doi.org/10.1145/3428226}
\showDOI{\tempurl}


\bibitem[Simonyan and Zisserman(2015)]%
        {vggsimonyan2015deep}
\bibfield{author}{\bibinfo{person}{Karen Simonyan} {and}
  \bibinfo{person}{Andrew Zisserman}.} \bibinfo{year}{2015}\natexlab{}.
\newblock \bibinfo{title}{Very Deep Convolutional Networks for Large-Scale
  Image Recognition}.
\newblock
\newblock
\showeprint[arxiv]{1409.1556}~[cs.CV]


\bibitem[Strout et~al\mbox{.}(2018)]%
        {polyinspect}
\bibfield{author}{\bibinfo{person}{Michelle~Mills Strout},
  \bibinfo{person}{Mary Hall}, {and} \bibinfo{person}{Catherine Olschanowsky}.}
  \bibinfo{year}{2018}\natexlab{}.
\newblock \showarticletitle{The Sparse Polyhedral Framework: Composing
  Compiler-Generated Inspector-Executor Code}.
\newblock \bibinfo{journal}{\emph{Proc. IEEE}} \bibinfo{volume}{106},
  \bibinfo{number}{11} (\bibinfo{year}{2018}), \bibinfo{pages}{1921--1934}.
\newblock
\urldef\tempurl%
\url{https://doi.org/10.1109/JPROC.2018.2857721}
\showDOI{\tempurl}


\bibitem[{V}an {Z}ee and {v}an~{d}e {G}eijn(2015)]%
        {BLIS}
\bibfield{author}{\bibinfo{person}{Field~G. {V}an {Z}ee} {and}
  \bibinfo{person}{Robert~A. {v}an~{d}e {G}eijn}.}
  \bibinfo{year}{2015}\natexlab{}.
\newblock \showarticletitle{{BLIS}: A Framework for Rapidly Instantiating
  {BLAS} Functionality}.
\newblock \bibinfo{journal}{\emph{ACM Trans. Math. Software}}
  \bibinfo{volume}{41}, \bibinfo{number}{3} (\bibinfo{date}{June}
  \bibinfo{year}{2015}), \bibinfo{pages}{14:1--14:33}.
\newblock
\urldef\tempurl%
\url{http://doi.acm.org/10.1145/2764454}
\showURL{%
\tempurl}


\bibitem[Wang et~al\mbox{.}(2021)]%
        {wang2021dualside}
\bibfield{author}{\bibinfo{person}{Yang Wang}, \bibinfo{person}{Chen Zhang},
  \bibinfo{person}{Zhiqiang Xie}, \bibinfo{person}{Cong Guo},
  \bibinfo{person}{Yunxin Liu}, {and} \bibinfo{person}{Jingwen Leng}.}
  \bibinfo{year}{2021}\natexlab{}.
\newblock \showarticletitle{Dual-Side Sparse Tensor Core}. In
  \bibinfo{booktitle}{\emph{Proceedings of the 48th {{Annual International
  Symposium}} on {{Computer Architecture}}}} \emph{(\bibinfo{series}{{{ISCA}}
  '21})}. \bibinfo{publisher}{{IEEE Press}}, \bibinfo{address}{{Virtual Event,
  Spain}}, \bibinfo{pages}{1083--1095}.
\newblock
\showISBNx{978-1-4503-9086-6}
\urldef\tempurl%
\url{https://doi.org/10.1109/ISCA52012.2021.00088}
\showDOI{\tempurl}


\bibitem[Wang(2021)]%
        {wang2021sparsednn}
\bibfield{author}{\bibinfo{person}{Ziheng Wang}.}
  \bibinfo{year}{2021}\natexlab{}.
\newblock \showarticletitle{{{SparseDNN}}: {{Fast Sparse Deep Learning
  Inference}} on {{CPUs}}}.
\newblock \bibinfo{journal}{\emph{arXiv:2101.07948 [cs]}} (\bibinfo{date}{July}
  \bibinfo{year}{2021}).
\newblock
\showeprint[arxiv]{2101.07948}~[cs]


\bibitem[Warden(2015)]%
        {warden2015why}
\bibfield{author}{\bibinfo{person}{Pete Warden}.}
  \bibinfo{year}{2015}\natexlab{}.
\newblock \bibinfo{title}{Why {{GEMM}} Is at the Heart of Deep Learning}.
\newblock
\newblock


\bibitem[Xianyi et~al\mbox{.}(2012)]%
        {xianyi2012openblas}
\bibfield{author}{\bibinfo{person}{Zhang Xianyi}, \bibinfo{person}{Wang Qian},
  {and} \bibinfo{person}{Zaheer Chothia}.} \bibinfo{year}{2012}\natexlab{}.
\newblock \showarticletitle{Openblas}.
\newblock \bibinfo{journal}{\emph{URL: \url{http://xianyi.github.io/OpenBLAS}}}
  (\bibinfo{year}{2012}), \bibinfo{pages}{88}.
\newblock


\bibitem[You et~al\mbox{.}(2022)]%
        {you2022drawing}
\bibfield{author}{\bibinfo{person}{Haoran You}, \bibinfo{person}{Chaojian Li},
  \bibinfo{person}{Pengfei Xu}, \bibinfo{person}{Yonggan Fu},
  \bibinfo{person}{Yue Wang}, \bibinfo{person}{Xiaohan Chen},
  \bibinfo{person}{Richard~G. Baraniuk}, \bibinfo{person}{Zhangyang Wang},
  {and} \bibinfo{person}{Yingyan Lin}.} \bibinfo{year}{2022}\natexlab{}.
\newblock \showarticletitle{Drawing {{Early-Bird Tickets}}: {{Towards More
  Efficient Training}} of {{Deep Networks}}}.
\newblock \bibinfo{journal}{\emph{arXiv:1909.11957 [cs, stat]}}
  (\bibinfo{date}{Feb.} \bibinfo{year}{2022}).
\newblock
\showeprint[arxiv]{1909.11957}~[cs, stat]


\bibitem[Yu et~al\mbox{.}(2017)]%
        {yu2017scalpel}
\bibfield{author}{\bibinfo{person}{Jiecao Yu}, \bibinfo{person}{Andrew
  Lukefahr}, \bibinfo{person}{David Palframan}, \bibinfo{person}{Ganesh
  Dasika}, \bibinfo{person}{Reetuparna Das}, {and} \bibinfo{person}{Scott
  Mahlke}.} \bibinfo{year}{2017}\natexlab{}.
\newblock \showarticletitle{Scalpel: {{Customizing DNN}} Pruning to the
  Underlying Hardware Parallelism}. In \bibinfo{booktitle}{\emph{2017
  {{ACM}}/{{IEEE}} 44th {{Annual International Symposium}} on {{Computer
  Architecture}} ({{ISCA}})}}. \bibinfo{pages}{548--560}.
\newblock
\urldef\tempurl%
\url{https://doi.org/10.1145/3079856.3080215}
\showDOI{\tempurl}


\bibitem[Zhang et~al\mbox{.}(2020)]%
        {zhang2020sparch}
\bibfield{author}{\bibinfo{person}{Zhekai Zhang}, \bibinfo{person}{Hanrui
  Wang}, \bibinfo{person}{Song Han}, {and} \bibinfo{person}{William~J. Dally}.}
  \bibinfo{year}{2020}\natexlab{}.
\newblock \showarticletitle{{{SpArch}}: {{Efficient Architecture}} for {{Sparse
  Matrix Multiplication}}}. In \bibinfo{booktitle}{\emph{2020 {{IEEE
  International Symposium}} on {{High Performance Computer Architecture}}
  ({{HPCA}})}}. \bibinfo{pages}{261--274}.
\newblock
\showISSN{2378-203X}
\urldef\tempurl%
\url{https://doi.org/10.1109/HPCA47549.2020.00030}
\showDOI{\tempurl}


\bibitem[Zhou et~al\mbox{.}(2021)]%
        {zhou2021learning}
\bibfield{author}{\bibinfo{person}{Aojun Zhou}, \bibinfo{person}{Yukun Ma},
  \bibinfo{person}{Junnan Zhu}, \bibinfo{person}{Jianbo Liu},
  \bibinfo{person}{Zhijie Zhang}, \bibinfo{person}{Kun Yuan},
  \bibinfo{person}{Wenxiu Sun}, {and} \bibinfo{person}{Hongsheng Li}.}
  \bibinfo{year}{2021}\natexlab{}.
\newblock \showarticletitle{Learning {{N}}:{{M Fine-grained Structured Sparse
  Neural Networks From Scratch}}}.
\newblock \bibinfo{journal}{\emph{arXiv:2102.04010 [cs]}}
  (\bibinfo{date}{April} \bibinfo{year}{2021}).
\newblock
\showeprint[arxiv]{2102.04010}~[cs]


\end{thebibliography}

\end{document}